# Artificial Intelligence in the Food Industry: Food Waste Estimation based on Computer Vision, a Brief Case Study in a University Dining Hall

*Shayan Rokhva, Babak Teimourpour\**

*Department of Information Technology, Faculty of Industrial and Systems Engineering, Tarbiat Modares University, Tehran, Iran*

[Shayanrokhva1999@gmail.com](Shayanrokhva1999@gmail.com) & [shayan_rokhva@modares.ac.ir](shayan_rokhva@modares.ac.ir)

[b.teimourpour@modares.ac.ir](b.teimourpour@modares.ac.ir) \*

## ABSTRACT:

Quantifying post-consumer food waste in institutional dining settings is essential for supporting data-driven sustainability strategies. This study presents a cost-effective computer vision framework that estimates plate-level food waste by utilizing semantic segmentation of RGB images taken before and after meal consumption across five Iranian dishes. Four fully supervised models—U-Net, U-Net++, and their lightweight variants—were trained using a capped dynamic inverse-frequency loss and AdamW optimizer, then evaluated through a comprehensive set of metrics, including Pixel Accuracy, Dice, IoU, and a custom-defined Distributional Pixel Agreement (DPA) metric tailored to the task. All models achieved satisfying performance, and for each food type, at least one model approached or surpassed 90% DPA, demonstrating strong alignment in pixel-wise proportion estimates. Lighter models with reduced parameter counts offered faster inference, achieving real-time throughput on an NVIDIA T4 GPU. Further analysis showed superior segmentation performance for dry and more rigid components (e.g., rice and fries), while more complex, fragmented, or viscous dishes, such as stews, showed reduced performance, specifically post-consumption. Despite limitations such as reliance on 2D imaging, constrained food variety, and manual data collection, the proposed framework is pioneering and represents a scalable, contactless solution for continuous monitoring of food consumption. This research lays foundational groundwork for automated, real-time waste tracking systems in large-scale food service environments and offers actionable insights and outlines feasible future directions for dining hall management and policymakers aiming to reduce institutional food waste.

# 1. Introduction
## 1.1. The Concept & Importance of the Study

Food waste has emerged as a pressing global issue, intersecting environmental, economic, and humanitarian concerns (Chawla and Lugosi, 2025; Sigala *et al.*, 2025). Extensive research in the food industry indicates that between one-fourth and one-third of all food produced is ultimately wasted, amounting to over one billion tons annually. This level of waste signifies a substantial depletion of vital resources, including water, land, labor, and energy. Beyond the ecological footprint, this inefficiency exacerbates food insecurity, especially in regions struggling with hunger and malnutrition (Attiq *et al.*, 2021; Gencia and Balan, 2024; Rokhva *et al.*, 2024a). Among the many sectors contributing to this challenge, institutional food services, particularly university dining halls, are often overlooked yet represent high-volume, high-variability consumption environments that lack fine-grained waste monitoring systems (Faezirad *et al.*, 2021; Leal Filho *et al.*, 2024). These settings present a unique opportunity to deploy scalable and intelligent monitoring tools to understand consumption patterns better and identify areas of inefficiency (Ahmadzadeh *et al.*, 2023).

Traditional approaches for quantifying food and food waste, such as physical weighing or subjective staff estimates, suffer from several limitations. They are not only time-consuming and prone to human error but also lack the granularity and objectivity required for longitudinal monitoring or behavioral analysis. Furthermore, they do not integrate seamlessly into modern digital infrastructures that are increasingly employed in smart kitchens and campus facilities (Ahmadzadeh *et al.*, 2023; Mazloumian *et al.*, 2020; Moumane *et al.*, 2023). As a result, there is growing interest in data-driven, automated approaches capable of producing reliable, scalable, and non-intrusive waste measurements. This is where artificial intelligence, and particularly computer vision, becomes highly relevant (Chakraborty and Aithal, 2024; Kumar *et al.*, 2022). Deep learning-based visual analysis methods offer significant potential in extracting meaningful insights from food tray images, enabling detailed evaluations of leftover food without interrupting the natural dining flow (Lubura *et al.*, 2022).

Recent advancements in computer vision have already transformed several domains. Although computer vision has seen transformative success in fields such as medical imaging and autonomous vehicles, its application in food-related domains, such as food recognition, portion estimation, and dietary assessment, has also advanced considerably in recent years (Aghajani *et al.*, 2024)bo, demonstrating the potential of visual models to extract meaningful nutritional and structural information from images (Khan *et al.*, 2021; Zhu *et al.*, 2021). Techniques such as convolutional neural networks (CNNs) and semantic segmentation have demonstrated the ability to discern food types, measure surface area, and even estimate caloric content from images (Amugongo *et al.*, 2023; Tao *et al.*, 2017). Yet, the specific task of estimating food waste by comparing the visual state of meals before and after consumption remains relatively underexplored. The primary technical challenge lies in the need for accurate, pixel-level delineation of food items, both in untouched and partially consumed states, which introduces variability due to occlusion, distortion, and lighting changes (Bohlol *et al.*, 2025). Nevertheless, semantic segmentation offers a viable pathway for quantifying such differences (Guerra Ibarra *et al.*, 2025), as surface-level changes can still serve as a reasonable proxy for estimating consumption, especially in structured and consistent environments like dining halls.

To address the lack of scalable, objective approaches for quantifying food waste in institutional environments, this study investigates the potential of computer vision to estimate food consumption by analyzing visual cues from structured dining scenarios. The primary objective is to explore whether surface-level visual analysis, derived from before- and after-meal imagery, can yield reliable estimations of food intake across a variety of food types. While this approach does not capture the volumetric or weight-based dimensions of waste, it offers a practical proxy through pixel-level surface comparison, particularly suitable for high-throughput, real-world settings such as university dining halls. By focusing on the visual differentiation of consumed versus remaining food portions, this work contributes to the evolving discourse on the role of AI in facilitating data-driven assessments of consumption behavior. Moreover, by grounding the study in operationally constrained environments and leveraging annotated visual datasets, the research aims to highlight both the opportunities and challenges associated with deploying intelligent analytics frameworks in institutional food service contexts.

## 1.2. Research Gap Identification

This section presents a comprehensive review of the most recent and relevant studies focusing on the application of AI and computer vision within the food industry, intending to identify and prove the specific research gap addressed in this work. To maintain clarity and avoid excessively lengthy textual descriptions, a structured summary is provided in **Table 1,** outlining the objectives, datasets, proposed methodologies, and, most critically, the advantages and limitations of each reviewed study.

Table 1. Comprehensive Review of the Most Recent & Relevant Studies Regarding the Application of AI and Computer Vision in the Food Industry and Food Science

| Reference | Novel/ Survey | Objective | Dataset | Proposed Method | Pros & Cons |
|---|---|---|---|---|---|
| (Ahmadzadeh et al., 2023) | Survey | Reviewing FWR[1] with IoT & Big Data | Not applicable | Not applicable | + Demonstrates AI & IoT benefits for FWR; Proposing General Framework <br> – Survey |
| (Mustapha et al., 2025) | Novel | Kitchen waste as compostable or not | 721 kitchen waste images | YOLOv8 + CNN: YOLOv8 detects ROIs → CNN classifies them | + Accurate real-time binary classification <br> – Small dataset; No quantity estimation; Binary only |
| (Wang et al., 2024) | Novel | Fruit & vegetable classification with ML & dimensionality reduction | Image-based from Kaggle | Dimensionality reduction + Classification (DT/SVM/etc) | + Diffusion maps with ML; High performance <br> – No timing, Limited to classification |
| (Vardopoulos et al., 2025) | Survey | Trends and gaps in food waste quantification | Not applicable | Not applicable | + Identification of trends & gaps in food industry <br> – Survey |
| (Thaseen Ikram et al., 2023) | Novel | Optimized IoT waste system via GA-Fuzzy | Simulated sensor data for waste levels in bins | IoT sensors → GA for route optimization → FIS for bin status decisions | + Combines IoT, optimization, fuzzy logic <br> – No visual or food-specific estimation; focuses on general bin fullness |
| (Fang et al., 2023) | Survey | Review of AI for smart city waste management | Not applicable | Not applicable | + Shows integration potential with AI, IoT, GIS, sensors <br> – Survey |
| (Abiyev and Adepoju, 2024) | Novel | Improved food recognition via DCNN + Attention | Food-101 & MA Food-121 | Custom CNN, Self-attention, Ensemble learning | + AI food recognition + diverse datasets; High performance <br> – Recognition only; No Volume estimation |
| (Rokhva et al., 2024a) | Novel | Speedy and accurate food recognition | Food11 | Pretrained MobileNetV2 + DA[2] + different image sizes | + Fast, accurate food recognition; Includes runtime <br> – Limited for multi-food dishes; No detection/segmentation |
| (Rokhva et al., 2024b) | Novel | Superior performance for food classification | Food 11 | EfficientNetB7 + DA + TL[3] | + Superior performance; State-of-the-art model <br> - Dense model & slow processing, Classification only |
| (Koirala et al., 2019) | Survey | Review for fruit detection & yield estimation | Not applicable | Not applicable | + Covers diverse DL methods, Yield estimation <br> – Survey |
| (Rokhva and Teimourpour, 2025) | Novel | Improved food recognition with still good speed | Food 11 | EfficientNetB7 + fine-tunning + CBAM attention | + Approaches ensemble method; maintains high speed <br> – Only 11 classes; limited adaptability; Recognition only |
| (Zhu et al., 2021) | Survey | DL and ML-vision in food processing | Not applicable | Not applicable | + Mentioning constraints, Review of ML-food processing <br> – Survey |
| (Park et al., 2021) | Novel | Real food segmentation via Sim-to-Real Mask R-CNN | Synthetic Blender data + real food dataset | Mask R-CNN trained on synthetic data; fine-tuned on real data for Sim-to-Real segmentation | + Effective Sim-to-Real segment; Use quality synthetic data <br> – No volume or mass estimation/quantification |
| (Mazloumian et al., 2020) | Novel | DL for food waste classification | 0.5M images; 1000 images with binary masks | U-Net for food region extraction → VGG16-based dual-path classifier using delta layers | + Delta-layer for temporal learning; before-after images <br> – No volume amount reported; segmentation used only in preprocessing; classifies only the last added item |
| (Geetha et al., 2022) | Novel | Waste management system ensemble neural network | Combined trash datasets; 6 classes; 2.5k images | Mask R-CNN + Ensemble classifiers + VGG16 + STL & SFM volume estimation | + Ensemble models; Volume consideration (with motion) <br> – Ensemble complexity; No runtime; Not food specifically |
| (Jubayer et al., 2021) | Novel | Mold detection using a YOLOv5-based DL model | 2050 mold-annotated images: 850 lab, 1200 web | YOLOv5 object detection + annotated mold images + DA | + Accurate detection aids FWR; biologically verified <br> – No mold type; No exact waste quantity |
| (Lubura et al., 2022) | Novel | Food recognition & rough waste estimation via CNNs | ~23.5K images (157-class) + ~1.35K-waste estimation (Serbian students) | Conventional CNN for classification, before and after masks for waste estimation | + Pioneering; Classification with real-meal dataset; rough waste estimation; Considering practical challenges <br> – No timing; No widely-used segmentation metrics |
| (Chen et al., 2025) | Novel | Food recognition via multi-level feature fusion and attention | CETH Food-101, Vireo Food-172, UEC Food-100 | Multi-stage feature fusion + self-attention + SCLoss → classification | + Attention for fine-grained recognition; Two benchmarks <br> – Lacks detection/segmentation; No thorough runtime |
| (Zhang et al., 2024) | Novel | Fruit freshness detection + multi-task CNN | 12,335-image; 8 fruit types, fresh/rotten labels | Shared CNN + depth-wise separable layers → MTL: freshness + fruit type | + Efficient MTL with shared features; accurate, lightweight; <br> – No timing metric; No freshness/ripeness/decomposition score or level |
| (Sigala et al., 2025) | Novel | Evaluating AI waste tracking in HORECA sites | Food waste data from five HORECA sites. | AI device, pre-post tracking, statistical & behavioral analysis | + Multiple kitchens; Hybrid estimation with camera & scale; <br> – Context sensitivity; Small sample |
| (Yuan and Chen, 2024) | Novel | Freshness detection via CNN features, PCA, & ML | 12,000 images (10 fruits + 10 vegetables, ~equal size | Different CNNs → Deep features fusion → PCA → Classifiers | + Pretrained models; PCA effective; Good performance <br> – Classification only; Limited for multi-object dishes |
| (Bohlol et al., 2025) | Novel | Recognizing 16 foods with optimized ResNet50 + custom FC layers | 66K augmented images; 16 food classes from meals under varying conditions | ResNet50 +DA + Fine-tuning + custom FC layers | + Custom FC layer; Comprehensive evaluation, Time report <br> _ Recognition only; Lower performance for some classes; No estimation/quantification/segmentation |
| (Razavi et al., 2024) | Novel | Rice classification and quality detection using ResNet + TL | 15625 augmented images (12500 original); 6 varieties | Differenr ResNets; DA + TL + Fine tunned FC layer → Classification | + Highly accurate in both classification and purity detection <br> – No segmentation or quantification; Limited use for FWR; No waste consideration |
| (Onyeaka et al., 2023) | Survey | AI role for FWR and circular economy | Not applicable | Not applicable | + Covers AI integration across food stages with examples <br> – Survey |
| (Shehzad et al., 2025) | Survey | Reviewing computer vision in food quality assessment | Not applicable | Not applicable | + Highlights practical computer vision use and emerging trends; validates research gaps <br> – Survey |

---

[1] Food Waste Reduction <br>
[2] Data Augmentation <br>
[3] Transfer Learning

A review of the most recent and relevant literature reveals that the application of computer vision for direct food waste estimation remains an underexplored research area. While numerous studies have effectively advanced food recognition, detection, and classification tasks, the specific goal of estimating food waste has received limited attention. Notably, the work by (Lubura *et al.*, 2022) presents a valuable early contribution by incorporating segmentation for waste assessment. However, the study does not employ predictive deep learning models such as U-Net or Nest-Net, and the methodology for quantifying food waste is only briefly addressed, with almost no use of typical segmentation metrics, leaving room for further development. This is particularly significant given that several recent studies, including both novel and survey-based research (Onyeaka *et al.*, 2023; Shehzad *et al.*, 2025; Sigala *et al.*, 2025; Vardopoulos *et al.*, 2025; Zhu *et al.*, 2021), have emphasized the potential of AI and computer vision in transforming food industry practices.

To bridge this gap, the present study employs computer vision techniques, specifically semantic segmentation, to compare food portions before and after consumption, thereby enabling the estimation and quantification of food waste solely based on image data. As with any research, this work may have certain limitations, which will be thoroughly discussed in the relevant section. Nonetheless, it proposes a novel and efficient framework for food waste estimation, grounded in predictive AI models that aim to deliver both accuracy and speed.

### 1.3. Research Efforts & Contributions

As outlined earlier, this study aims to quantify food waste percentages through computer vision by analyzing and comparing food images captured before and after consumption, an area that remains underexplored and represents a notable research gap. Accordingly, the primary contributions of this work are summarized as follows:

I. Development of a segmentation dataset from a university dining hall, including image collection, manual annotation of masks, and required augmentations. The process also involves identifying dataset-specific challenges, selecting appropriate evaluation metrics, designing customized metrics, and providing an in-depth discussion and limitations, beneficial for future research in this domain.
II. Implementation of food waste and consumption rate estimation by analyzing annotated segmentation masks of food dishes before and after consumption.
III. Proposal of four semantic segmentation models, including standard U-Net and Nested U-Net for high-accuracy segmentation, as well as their lightweight variants to balance performance with faster inference times.
IV. Empirical evaluation of the proposed predictive models against manual estimations, assessing their performance and the study's effectiveness, while also outlining current considerations and opportunities for further improvements.

The remainder of this study is organized as follows. Section 2 outlines the materials and methods, providing comprehensive details on the dataset, the proposed methodology, and the metrics utilized and defined. Section 3 presents the results along with a thorough analysis. Section 4 offers an in-depth discussion concerning the effectiveness of the employed methodology, the study's limitations and potential biases, real-world applications, and opportunities for future improvements. Finally, Section 5 concludes the paper.

## 2. Materials & Methods

This section outlines the materials and methods, beginning with dataset preparation, covering data collection, annotation, and augmentation. It then describes the proposed approach, which involves estimating food waste using annotated masks, applying predictive segmentation models, and model details. Finally, it delves into the evaluation metrics, including both standard and customized measures. This comprehensive explanation facilitates reproducibility and potential enhancement in future research.

### 2.1. Dataset
#### 2.1.1. Data Curation

The dataset used in this study comprises five categories of cooked meals served at the dining hall of a major public university in Tehran, Iran. Although the university dining hall offers a diverse menu, providing approximately 15 different meal types over a two-week period, only five of the most frequent, popular, and widely consumed meals were selected due to the time-consuming and demanding nature of generating precise masks for segmentation tasks. These categories include *Adas Polo*, *Chelo-Goosht*, *Fesenjan Stew with Rice*, *Gheymeh Bademjan Stew with Rice*, and *Protein with French Fries* (which encompasses items such as chicken schnitzel, cordon bleu, and chicken nuggets). For better comprehension, a summary of each food category, including its Persian name, English equivalent, and key ingredients, is provided in **Table 2**. These English names are approximate translations aimed at improving clarity, not exact culinary counterparts.

For each category, images of the plates were captured by the authors in two states: before and after consumption. Considering the high volume of meals served daily, and consequently, the potentially large number of images, and the associated challenges of generating accurate masks, a limited (yet sufficient for training segmentation models) number of images was selected for each food category. It is therefore important to emphasize that the results of this study are intended solely to introduce and evaluate a computer vision-based framework for estimating food waste, and may not provide an exact statistical estimation of food waste levels in the university dining hall. Naturally, this framework holds significant potential for scaling up; if supported by adequate financial and research resources, the very same methodology can be applied to a substantially larger dataset with corresponding masks to further enhance segmentation model performance.

Regarding the imaging process, photos were taken by the authors using a SAMSUNG A54 smartphone to ensure high-quality image resolution, minimizing quality degradation during future resizing. Given that the relative distance of the camera from the food tray may introduce bias into the data, care was taken to capture all images from a reasonably consistent and appropriate distance to reduce any data collection bias. Admittedly, this constraint stems from current research limitations and can be effectively addressed in future work by employing a fixed camera setup installed above the serving trays.

Additionally, it is important to note that the images collected for each food category, before and after consumption, are neither equal in number nor strictly paired (i.e., image pairs of the same dish before and after consumption). This characteristic reflects an attempt to simulate real-world conditions within a university dining environment, as well as a reasonable simplification. In such settings, food is typically served in relatively equal portions to all students to maintain fairness. As such, the distribution of food pixels in the "before consumption" state is expected to vary within a narrow, consistent range for each food category. Accordingly, by calculating the ratio between the post-consumption remaining food pixels and the average pre-consumption food pixels, a logical estimate of the consumed and wasted portions can be derived. Moreover, it is practically infeasible to locate and track the exact plate served to a particular student in a large-scale dining hall.

Finally, to aid comprehension, **Figure 1** illustrates a subset of images from the five food categories in both pre- and post-consumption states, validating the above-stated considerations. As shown, the images were captured under varying lighting conditions to better simulate real-world scenarios. Furthermore, the forthcoming discussion on data augmentations and transformations confirms that the diversity introduced at this stage enhances the model's learning capability and strengthens its applicability to real-world conditions.

Table 2. Overview of Persian Dishes and Their English Descriptions

| Persian food name | English food name + Ingredients |
|---|---|
| AdasPolo | Lentil Rice with Raisins and Meat |
| CheloGoosht | Rice with Beef |
| Fesenjan | Pomegranate-Walnut Stew with Rice |
| GheymeBademjan | Persian Split Pea & Eggplant Stew with Rice |
| Protein & Fries | Chicken schnitzel, Cordon bleu, and Chicken nuggets with French fries |

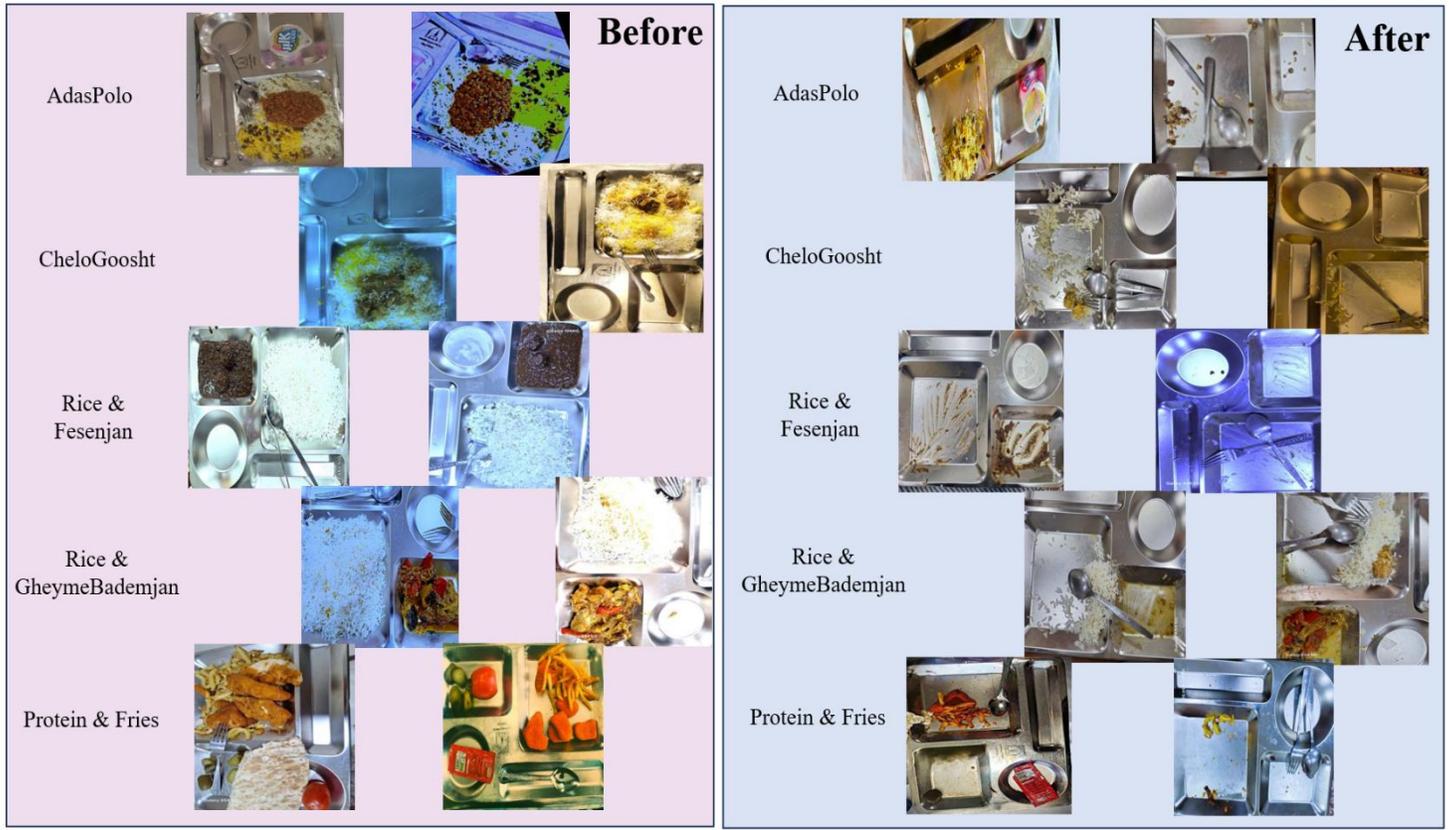

Figure 1. Representative Pre- & Post-Consumption Images for Each Food Category

### 2.1.2. Annotation and Semantic Mask Generation

Following the collection of images across the five selected food categories, it was necessary to generate semantic masks for each image to prepare the dataset for further processing. To this end, the Roboflow platform was utilized to perform detailed labeling and semantic segmentation of the key components of each dish. In all categories, class zero was assigned to the background, while the remaining classes were defined based on the main components of each specific food item. To provide an overview, **Table 3** presents the number and types of segmentation classes for each food category.

Table 3. Class Numbers and Annotations by Food Type

| Food Type | Class Numbers | Classes + (0: background) |
|---|---|---|
| AdasPolo | 2 | 1: AdasPolo |
| CheloGoosht | 3 | 1: Meat<br>2: Rice |
| Fesenjan | 3 | 1: Fesenjan stew<br>2: Rice |
| GheymeBademjan | 3 | 1: GheymeBademjan stew<br>2: Rice |
| Protein & Fries | 3 | 1: French fries<br>2: Protein |

All images, both before and after consumption, were subjected to semantic segmentation. Typically, images taken before consumption contain all defined classes, including both background and main food components. However, in the post-consumption images, one or more primary food components may be entirely absent or present in significantly reduced quantities. To better illustrate this, **Figure 2** displays sample images captured before and after consumption alongside their corresponding segmentation masks. Distinctive colors have been employed in the masks to facilitate clearer visual differentiation of each class.

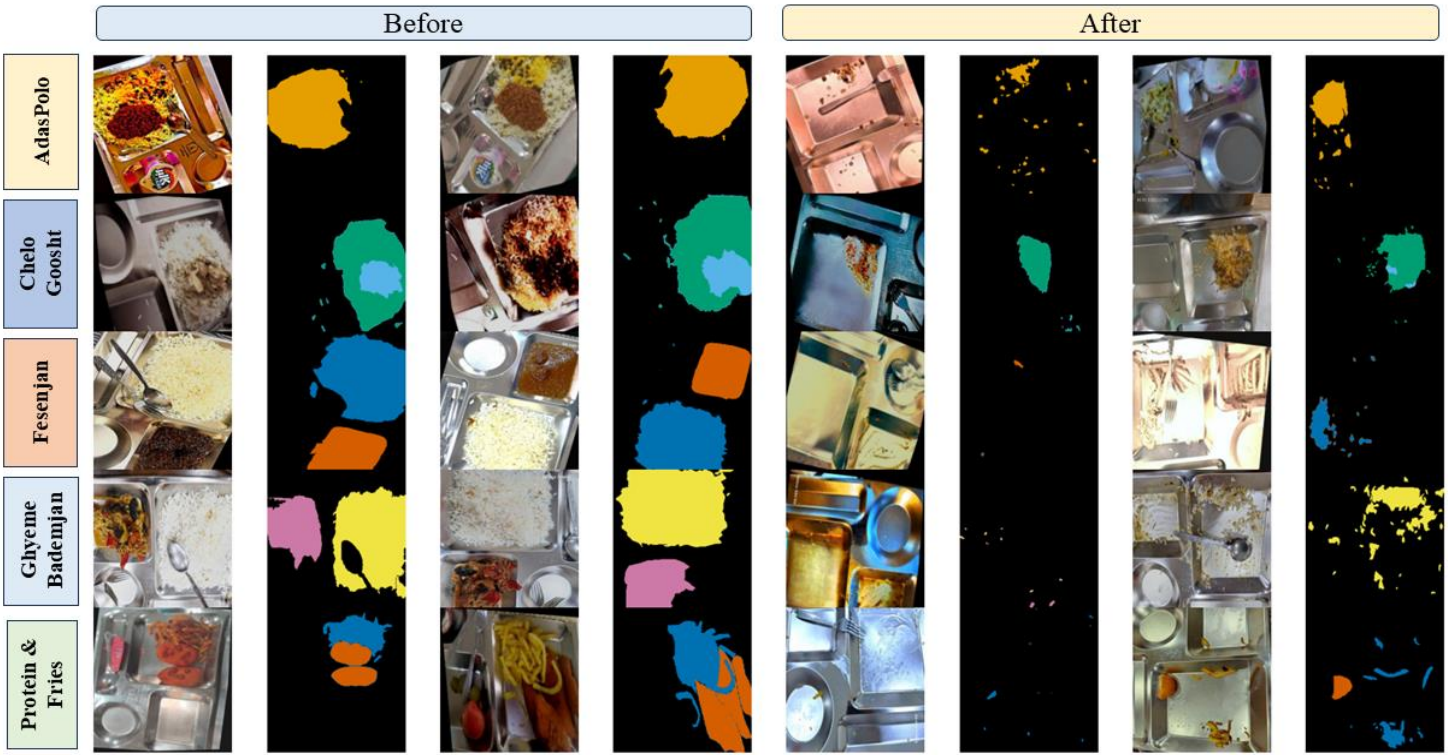

Figure 2. Sample Images & Segmentation Masks Before & After Consumption Across Food Categories

### 2.1.3. Transformation & Augmentation

To prepare the datasets for the five food categories, all images were resized to 256×256 pixels. The training and test sets were then separated at the beginning to prevent data leakage and minimize potential bias. It is worth noting that, in many cases, strong visual similarity may exist among pre-consumption dishes or nearly empty post-consumption plates. This similarity is inherent to the problem context and should not be interpreted as a sign of bias or leakage, since, as stated, the training and test data for each food category were separated from the beginning. The test set proportion varied across categories but, on average, constituted approximately 20% to 30% of the total data, based on the available dataset size.

Following the exclusion of test data, data augmentation was applied to the training set to triple its original size. This process was facilitated using the free tier of the Roboflow platform. The augmentations were defined randomly by the authors and, while they may slightly differ between food categories, they were designed comprehensively enough to reflect a wide range of real-world conditions, such as changes in lighting, sharpness, viewing angle, image flipping, and more. For more detail, **Table 4** presents the set of random augmentations applied to the five food categories as well as the number of images in the train and test subsets.

It is important to note that the augmentations listed in **Table 4** are probabilistically applied; hence, any given image may undergo none, one, several, or even all of the listed transformations. However, due to the sequential and probabilistic structure of the pipeline, the likelihood of an image experiencing the full set of augmentations is very low. As a result, the augmented images remain realistic and free from excessive distortions that could impair model generalization. To further clarify how these augmentations affect the visual properties of the images, **Figure 3** illustrates an example image subjected to several possible transformations. Each instance in the figure represents the application of a single augmentation to the original image.

Table 4. Augmentation Pipeline & Dataset Split Statistics

| Food Type | Augmentation for Training Set (3x) | | | | | | | | | Test size | Train size pre-augment |
|---|---|---|---|---|---|---|---|---|---|---|---|
| | Step 1 | Step 2 | Step 3 | Step 4 | Step 5 | Step 6 | Step 7 | Step 8 | Step 9 | | |
| **Adas Polo** | Flip (H/V) | 90-Rotate (CW/CCW/180) | Rotation (-15 to +15) | Shear (+-10 H&V) | Saturation (-5 to +5) | Brightness (-10 to +10) | Exposure (-3 to +3) | Blur (Up to 1px) | None | 63 | 264 |
| **Chelo Goosht** | Flip (H/V) | 90-Rotate (CW/CCW/180) | Rotation (-15 to +15) | Shear (+-15 H&V) | Hue (-15 to +15) | Saturation (-20 to +20) | Brightness (-15 to +15) | Exposure (-5 to +5) | None | 64 | 207 |
| **Fesenjan** | Flip (H/V) | 90-Rotate (CW/CCW/180) | Rotation (-15 to +15) | Shear (+-15 H&V) | Hue (-18 to +18) | Saturation (-15 to +15) | Brightness (-10 to +10) | Exposure (-10 to +10) | Blur (Up to 1.3px) | 61 | 148 |
| **Gheyme Bademjan** | Flip (H/V) | 90-Rotate (CW/CCW/180) | Rotation (-15 to +15) | Shear (+-15 H&V) | Hue (-15 to +15) | Saturation (-15 to +15) | Brightness (-15 to +15) | Blur (Up to 1.2px) | None | 63 | 167 |
| **Protein & Fries** | Flip (H/V) | 90-Rotate (CW/CCW/180) | Rotation (-15 to +15) | Shear (+-15 H&V) | Hue (-15 to +15) | Saturation (-10 to +10) | Brightness (-15 to +15) | Exposure (-5 to +5) | Blur (Up to 1.2px) | 103 | 273 |

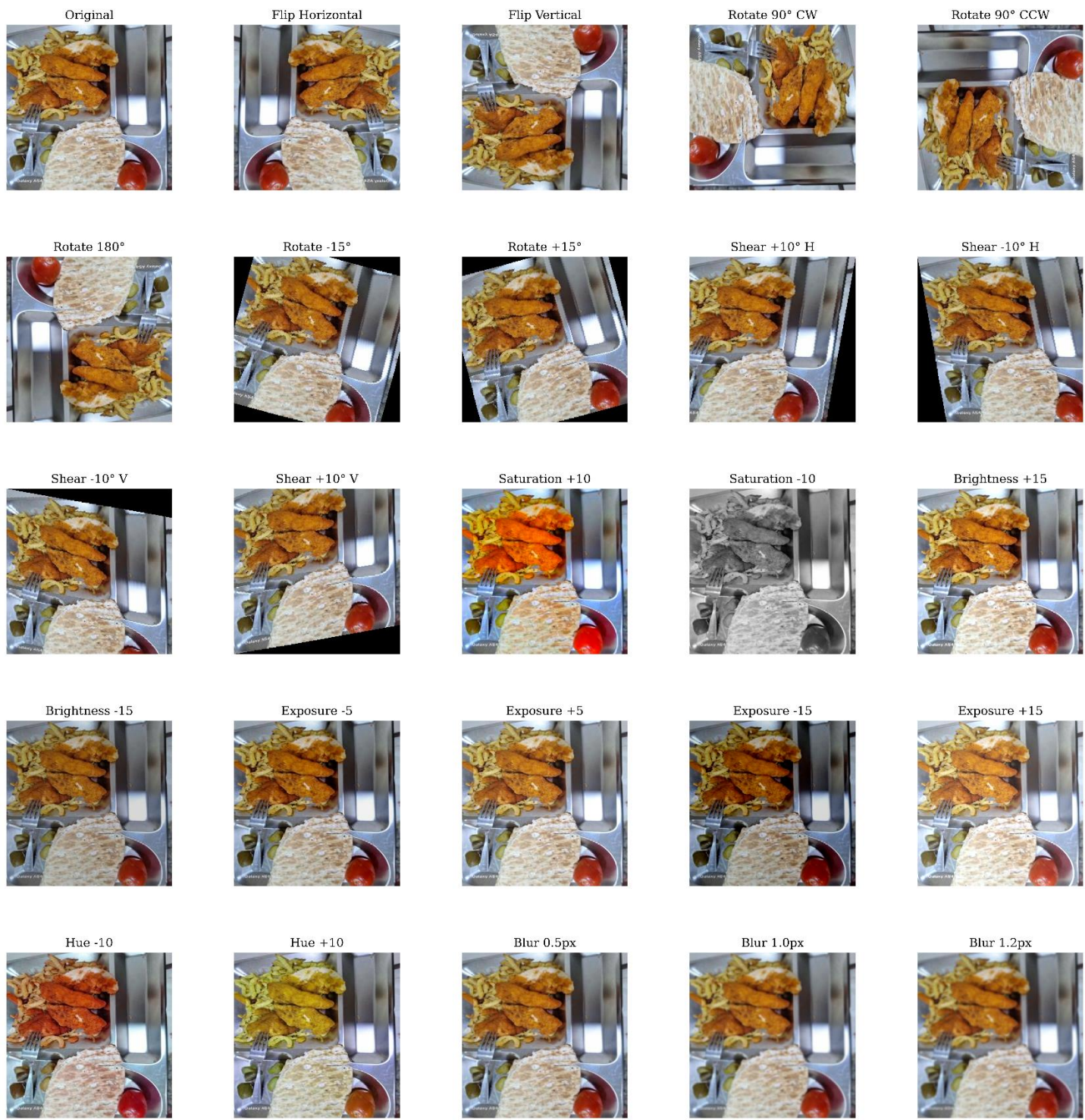

Figure 3. Several Single-Step Augmentations Applied to a Sample Image

## 2.2. Proposed Method

This subsection presents a detailed overview of the methodology adopted in this study. Initially, consumption and remaining rates are derived by comparing segmentation masks of dishes captured before and after consumption. Subsequently, four predictive models, inspired by U-Net and U-Net++ architectures, are proposed to estimate the performance of segmentation and, subsequently, the effectiveness of automated food waste estimation via the proposed models, without the need for manual mask segmentation every time. The final part addresses the model configurations that influenced the observed model performance.

### 2.2.1. Actual Food Waste Estimation

The main objective of this phase is to estimate the food consumption and unconsumed rate by analyzing the areas covered by individual food components and the background in each image. By computing the proportion of each component from the segmentation masks, the food waste rate, defined as the complement of the consumption rate, can be accurately determined. Given that these masks were manually created by the authors, they are assumed to be precise and thus serve as a robust ground truth for training AI models and evaluating segmentation performance, eliminating the need for repeated manual labeling.

Assuming a dish contains **n** classes indexed from **0** to **n−1**, and given 256×256-pixel input images, the proportion of class $i \in \{0, \ldots, n-1\}$ is calculated as shown in **Equation (1)**. Utilizing that, the proportion of each class is computed for all images in a given food category, both before and after consumption. Furthermore, these numbers can be averaged across all dishes to provide overall figures.

$$\text{Class i proportion} = \left(\frac{\text{Total pixels of class i}}{256 \times 256}\right) \times 100 \tag{1}$$

In the pre-consumption state, food portions served to students are generally within a specific range, with some variation in spatial distribution on the plate. As such, the distribution of each class is expected to fall within a rational range. Nevertheless, with or without this assumption, the weighted average class proportion in the pre-consumption state can be used as a benchmark, as done in this study. Hence, the eating rate for a specific class within a certain dish is then obtained by comparing its post-consumption area to this average pre-consumption benchmark. Based on this comparison, the eating rate for each class, excluding the background, is derived using **Equation (2).**

$$\text{Eating Rate}_i = \left(\frac{\overline{P}_i^{\text{pre}} - P_i^{\text{post}}}{\overline{P}_i^{\text{pre}}}\right) \times 100 \tag{2}$$

After calculating the consumption rate of a specific class within an individual dish, its average remaining rate is determined by averaging these values across all relevant dishes (post-consumption dishes) of that food type, as expressed in **Equation (3).** In this equation, **N** represents the number of images/masks.

$$\overline{\text{Eating Rate}}_i = \left(\frac{1}{N}\sum_{j=1}^{N} \text{Eating Rate}_i^{(j)}\right) \tag{3}$$

Since the remaining rate is the direct complement of the eating rate, the average remaining rate can be computed accordingly, as defined in **Equation (4).**

$$\overline{Remaining\ Rate}_i = 100 - \overline{Eating\ Rate}_i \tag{4}$$

### 2.2.2. Predictive Models

In this section, we introduce four segmentation models inspired by the U-Net and U-Net++ architectures, aiming to predict segmentation masks effectively. These models facilitate accurate predictions while minimizing the reliance on manual annotation, thereby automating the process of food waste estimation. As a baseline, we employ the original U-Net architecture proposed by (Ronneberger *et al.*, 2015) as is widely recognized for its simplicity, interpretability, and efficiency.

In the next step, to address the computational demands of real-world food segmentation, with rapid processing and a reduced number of parameters, we also developed a lightweight variant of U-Net. This modified version retains the original structural framework but reduces the number of 2D convolutional filters across each block. Specifically, whereas the standard U-Net includes 64, 128, 256, 512, and 1024 filters at each block of increasing depths, the lightweight version employs 32, 64, 128, 256, and 512 filters, respectively. These architectural changes, illustrated in **Figure 4** (Klingler, 2024)**,** lead to a decrease in model parameters and an improvement in processing speed. However, the reduced representational capacity of the lightweight model may limit its performance in more complex segmentation tasks, an aspect further examined in the results and analysis.

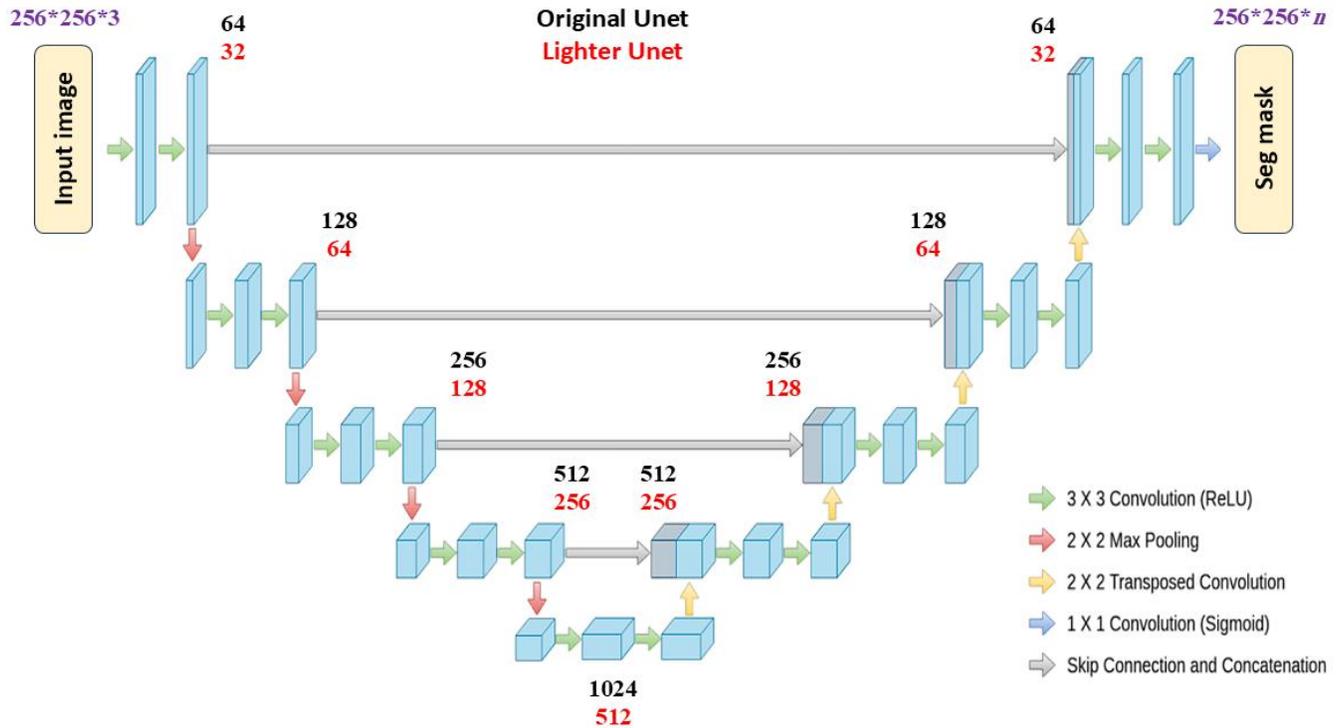

Figure 4. U-net VS Constructed Lighter U-net

In addition to the standard U-Net, U-Net++, also referred to as NestNet (Zhou *et al.*, 2018), is incorporated due to its ability to better preserve segmentation details through nested skip connections. Although these nested structures may introduce additional complexity in hyperparameter tuning, they may (or may not) offer improved performance in more challenging segmentation tasks. Therefore, both the original U-Net++ ("Papers with Code - UNet++ Explained", n.d.) and a lightweight variant with a reduced structure, as shown in **Figure 5,** are proposed in this study. Moreover, to enable a comparison of model complexity, **Table 5** presents the number of parameters for each proposed model along with the corresponding tensor shape at the bottleneck layer.

Table 5. Comparative Analysis of Model Complexity and Bottleneck Feature Dimensions

| Model | Param Counts | Bottleneck Tensor Shape |
|---|---|---|
| Unet | 31.04 m | [16*16*1024] |
| Smaller Unet | 7.77 m | [16*16*512] |
| NestNet (Unet++) | 36.15 m | [16*16*1024] |
| Smaller NestNet (Smaller Unet++) | 9.04 m | [16*16*512] |

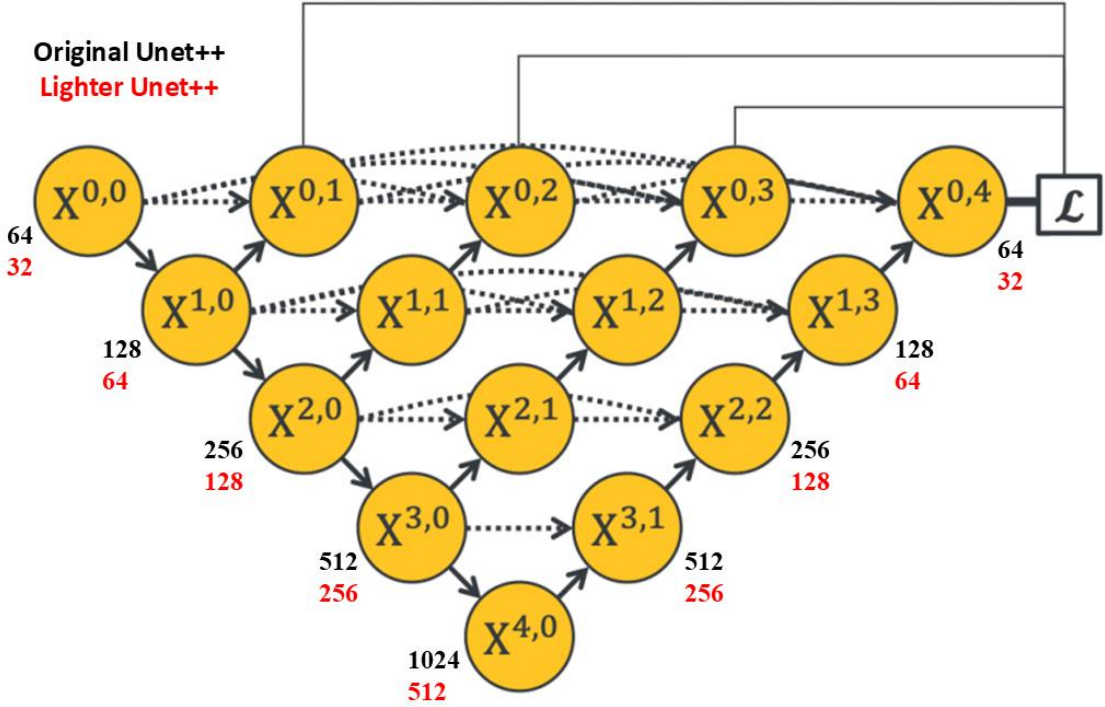

Figure 5. Original VS Constructed Lighter Unet++

### 2.2.3. Pipeline Components
#### a) Capped Dynamic Inverse-Frequency Cross Entropy Loss

To address the challenge of class imbalance in semantic segmentation tasks, we propose a refined dynamic weighting strategy integrated into the cross-entropy loss function. While standard cross-entropy assigns equal contribution to all semantic classes, this assumption fails in contexts where the pixel distribution is heavily skewed (e.g., large dominance of background or rice classes in food segmentation). In such cases, minority classes may be underrepresented, leading to suboptimal learning and poor segmentation fidelity. Our solution builds upon the dynamic class weighting strategy by computing per-batch weights inversely proportional to the prevalence of each class. For a given batch, let $f_c$ denote the number of pixels belonging to class $c$, and $C$ the total number of classes. The raw inverse-frequency weight for class ccc is computed in **Equation (5)**:

$$w_c = \frac{1}{f_c + \varepsilon} \tag{5}$$

To ensure numerical stability, a small constant $\varepsilon$ is added. While this approach effectively emphasizes rare classes, it may lead to excessive penalization when certain classes appear only in trace amounts (e.g., a single or a few protein pixels). To mitigate this risk, we introduce a capping mechanism to constrain the dynamic range of weights, as presented in **Equation (6).**

$$w_c^{capped} = min\left(w_c,\ \min_j(w_j) \cdot \lambda\right) \tag{6}$$

Where $\lambda \in \mathbf{R}^+$ is a hyperparameter denoting the maximum allowed ratio between the highest and lowest weights (we use $\lambda = 10.0$ in our experiments). This ensures that extremely rare classes do not disproportionately dominate the loss signal, while still allowing meaningful prioritization.

After applying the cap, weights are normalized to maintain a balanced aggregate contribution across classes, as presented in **Equation (7):**

$$\widehat{w_c} = \frac{w_c^{capped}}{\sum_{j=1}^{C} w_j^{capped}} \cdot C \tag{7}$$

The final loss is computed as a weighted pixel-wise cross-entropy, presented in **Equation (8)**:

$$\mathcal{L} = -\frac{1}{N} \sum_{i=1}^{N} \widehat{w_{y_i}} \cdot \log \widehat{p_{y_i}} \tag{8}$$

Where $N$ is the total number of pixels, $y_i \in \{0, 1, \dots, C-1\}$, is the ground truth label of pixel $i$, and $\widehat{p_{y_i}}$ is the predicted Softmax probability for the true class.

This capped inverse-frequency formulation strikes a balance between minority class sensitivity and loss stability, preventing rare classes from dominating training while still ensuring they are not ignored. It is particularly well-suited for real-world applications, such as food segmentation, where the semantic importance of smaller regions must be acknowledged without overwhelming the model's training dynamics.

**b) Adam VS AdamW**

The initial choice for optimization in this study was the Adam optimizer, due to its ability to adaptively adjust learning rates for each parameter and its momentum-like behavior via moving averages of gradients. These features allow Adam to achieve faster convergence and make it less sensitive to hyperparameter tuning compared to traditional optimizers like SGD (Kingma and Ba, 2017). However, as our objective also included applying L2 regularization to mitigate overfitting, the standard Adam optimizer presented limitations. In Adam, the weight update rule is shown in **Equation (9)**:

$$\theta_{t+1} = \theta_t - \eta \cdot \frac{m_t}{\sqrt{v_t} + \epsilon} \tag{9}$$

When L2 regularization (weight decay) is applied in this formulation, it is coupled with the gradient and thus affected by the adaptive learning rates. This causes the regularization effect to become inconsistent and less effective across parameters, especially in deep architectures where parameter-wise variances vary significantly. To address this, we adopted the AdamW optimizer, which decouples the weight decay term from the gradient update and applies it directly to the weights (Loshchilov and Hutter, 2019). The update rule for AdamW is presented in **Equation (10)**:

$$\theta_{t+1} = \theta_t \cdot (1 - \eta \cdot \lambda) - \eta \cdot \frac{m_t}{\sqrt{v_t} + \epsilon} \tag{10}$$

This decoupling ensures that weight decay behaves as intended, independent of the adaptive gradient scaling, leading to more stable training and improved generalization. Therefore, AdamW was selected as the final optimizer in this study to combine the benefits of fast convergence with effective regularization.

**c) Hyperparameter Values**

Properly selected hyperparameters are crucial for achieving stable convergence, efficient model training, and results near the optimal solution. As such, they must be carefully tuned and clearly reported.

**Hardware Configuration:** Although hardware specifications are not categorized as hyperparameters, reporting them remains essential due to their substantial influence on both training and inference efficiency. Training deep learning models of this scale is computationally infeasible on CPUs alone. Therefore, all experiments in this study were conducted using the NVIDIA T4 GPU provided via Google Colab, which significantly accelerated the training process and ensured timely convergence.

**Image size:** The input image resolution employed in this study is 256×256, a commonly adopted dimension across tasks such as classification, object detection, and semantic segmentation. While increasing image resolution can enhance model performance by preserving more spatial details—potentially with different hyperparameter configurations—it also introduces higher computational costs and slower inference. Conversely, reducing image size improves computational efficiency and speed but can severely degrade performance due to loss of critical visual details and context (Rokhva *et al.*, 2024a; Rukundo, 2023). Thus, a resolution of 256 was selected as a balanced trade-off between performance and computational efficiency.

**Batch Size:** A batch size of 4 was utilized consistently across training, validation, and testing phases. Unlike classification tasks, where batch sizes of 16, 32, or higher are common, semantic segmentation involves dense pixel-wise predictions and thus demands significantly more memory per sample. This relatively small batch size was chosen to optimize GPU memory usage while maintaining gradient stability and effective convergence.

**Learning Rate:** The learning rate is widely regarded as one of the most critical hyperparameters in the training of deep learning models, as it directly governs the optimization dynamics. If set too high, the model may diverge and fail to converge altogether; conversely, excessively small values can lead to prohibitively slow training or stagnation. Therefore, its selection demands careful empirical calibration. To identify a suitable range, preliminary experiments were conducted using values of 0.1, 0.01, 0.001, and 0.0001. In the context of this study and using the AdamW optimizer, a learning rate of 0.001 yielded the best (the fastest despite some minor fluctuations) initial convergence behavior, whereas larger values (0.1 and 0.01) introduced instability and significant loss oscillations. A rate of 0.0001 also produced stable and effective results, but at the expense of slower convergence, which is not suitable in the early epochs.

To have an effective learning dynamics throughout training, we implemented a progressive learning rate scheduler. While AdamW inherently adapts learning rates per parameter and thus reduces reliance on explicit schedulers (especially in comparison to SGD), incorporating a scheduler still offers significant benefits, particularly by enabling faster learning in early stages and finer adjustments in later epochs. As visualized in **Figure 6**, this tiered decay strategy spans multiple orders of magnitude, offering a robust mechanism for stable and efficient convergence. Even in instances where the initial learning rate (e.g., 1e-3) may be slightly aggressive (in some cases), the subsequent reductions act as a compensatory mechanism, promoting long-term stability. Similarly, later values such as 1e-5, even if they are found relatively small, do not hinder convergence but rather assist in fine-tuning. Altogether, this dynamic scheduling approach ensures both early optimization momentum and effective convergence toward minima.

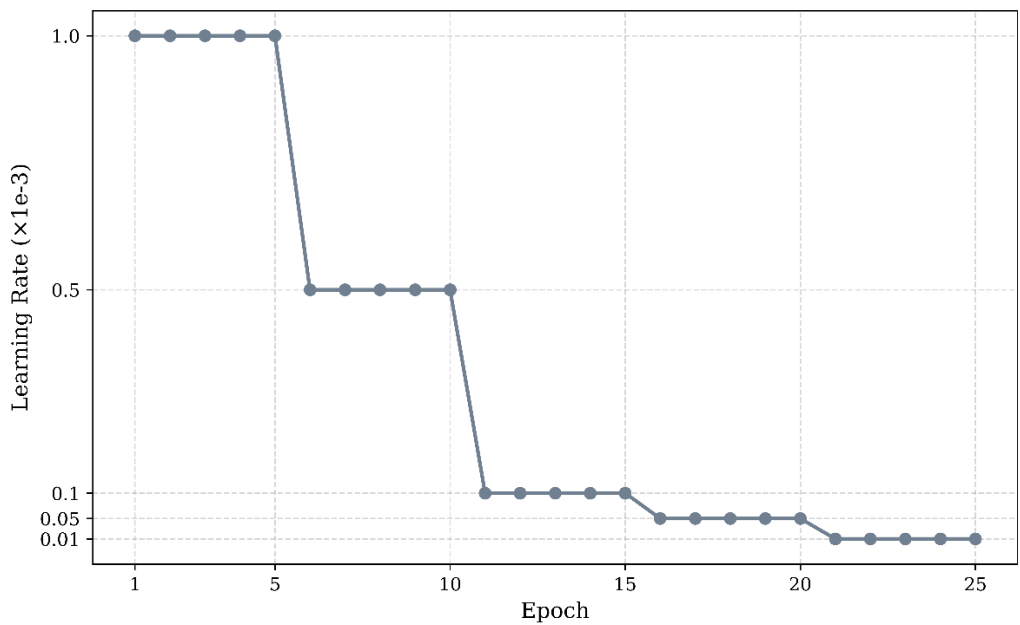

Figure 6. Progressive Learning Rate Adjustment Strategy Used During Training

**Regularization:** L2 Regularization is implemented via the "weight_decay" parameter within the optimizer (Pytorch implementation). It serves as a critical mechanism for controlling overfitting by penalizing large weights and stabilizing the training process. Given the dataset size, input image resolution, and model complexity in this study, a weight decay value of 1e-4 was empirically chosen. This setting should mitigate performance fluctuations while avoiding overfitting.

## 2.3. Evaluation Metrics
### 2.3.1. Predefined Metrics

To evaluate segmentation performance, we utilize three widely accepted predefined metrics. These metrics are commonly used in semantic segmentation tasks due to their ability to assess both overall correctness and class-wise overlap.

**Pixel Accuracy:** As presented in **Equation (11)**, it measures the ratio of correctly classified pixels to the total number of pixels. Importantly, this metric can be inflated by large/background classes, requiring the use of other segmentation metrics, especially in sensitive tasks.

$$\text{Pixel Accuracy} = \frac{\sum_{c=0}^{C-1} TP_c}{\sum_{c=0}^{C-1}(TP_c + FP_c + FN_c)} \quad (11)$$

**IoU:** Intersection over Union (IoU) measures the overlap between the predicted and ground truth segmentation for each class, with the formula shown in **Equation (12):**

$$\text{IoU}_c = \frac{TP_c}{TP_c + FP_c + FN_c} \quad (12)$$

**Dice Score:** A harmonic mean of precision and recall, focusing on overlap between prediction and ground truth. The formulation is provided in **Equation (13):**

$$\text{Dice}_c = \frac{2 \cdot TP_c}{2 \cdot TP_c + FP_c + FN_c} \quad (13)$$

These metrics together ensure a comprehensive evaluation: pixel accuracy gives a general overview, while IoU and Dice provide robust insight into class-wise performance, particularly useful when dealing with underrepresented classes.

### 2.3.2. Distributional Pixel Accuracy (DPA)

In addition to employing standard evaluation metrics, we introduce a novel metric termed Distributional Pixel Accuracy (DPA), which serves as a more lenient variant of traditional pixel accuracy, tailored to the unique demands of our application. Specifically, in the context of food waste estimation, the goal is not to precisely localize individual pixels but rather to estimate the proportions (proportional distribution) of different classes within each segmentation mask! To address this, DPA, as formulated in **Equation (14),** evaluates the class-wise pixel ratio similarity between the ground truth and predicted masks, emphasizing distributional agreement rather than spatial alignment.

$$\text{DPA}_c = 1 - \left| \frac{P_c^{\text{pred}}}{T^{\text{pred}}} - \frac{P_c^{\text{gt}}}{T^{\text{gt}}} \right| \quad (14)$$

Where:

- $P_c^{\text{pred}}$ and $P_c^{\text{gt}}$ denote the number of pixels predicted and labeled as class *c*, respectively.
- $T^{\text{pred}}$ and $T^{\text{gt}}$ are the total number of pixels in the predicted and ground truth masks, respectively.

The metric penalizes large deviations in the class proportion, thereby offering a distribution-focused performance measure that aligns with estimation-based goals.

To provide a clearer understanding, **Figure 7** illustrates some examples, including an extreme case in which conventional metrics (Pixel Accuracy, IoU, and Dice Score) yield values close to zero due to no spatial alignment, while the DPA metric approaches 1 as it correctly captures the class-wise distributional proportions! This example, while extremely rare, highlights the unconventional, yet beneficial, nature of DPA.

It is also important to note that DPA is not suitable for applications requiring high spatial precision, such as medical imaging, autonomous driving, or any other safety-critical or sensitive systems. However, in the context of food waste estimation, where the quantitative coverage of food items is more relevant than their precise localization, DPA proves useful. It enables proportional surface comparisons across classes, offering a practical perspective for evaluating waste estimation tasks.

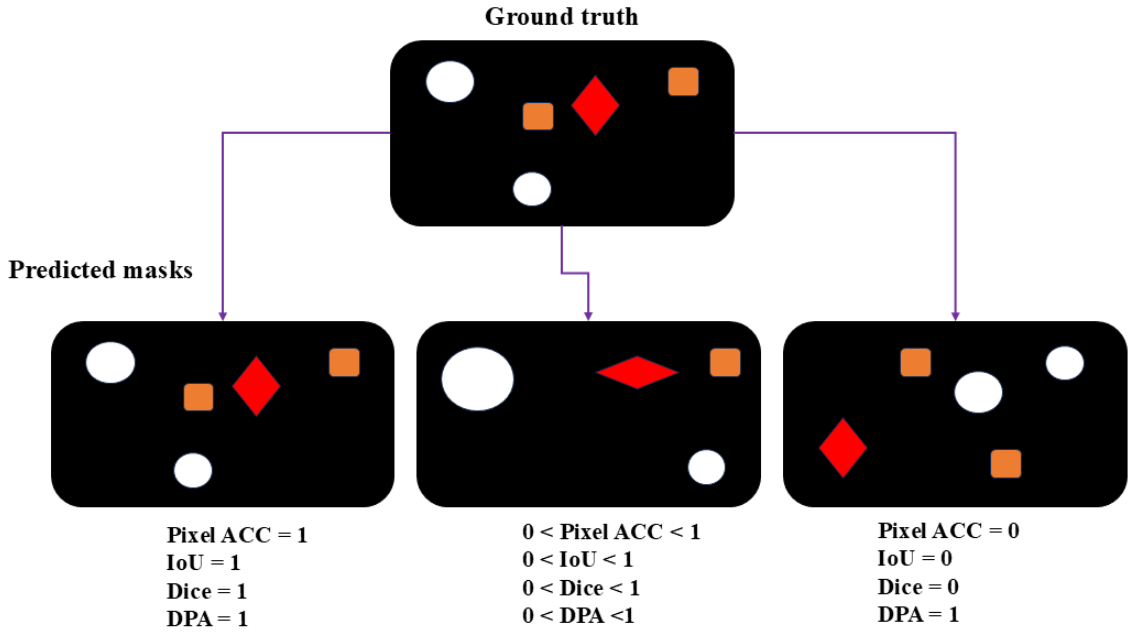

Figure 7. Illustration of DPA Compared to Traditional Metrics Under Varying Spatial Alignments

### 2.3.3. Macro VS Weighted

Selecting appropriate evaluation metrics, whether standard or tailored to the task, is essential for accurately measuring model performance. Macro averaging treats each class equally by computing metrics independently for each class and then taking a simple average, making it more responsive to the performance of minority classes. Weighted averaging, on the other hand, adjusts for class imbalance by computing per-class metrics and averaging them in proportion to the number of true instances per class. This leads to a more representative overall score, as dominant classes contribute more heavily to the final evaluation.

The decision between macro and weighted metrics should reflect the specific requirements of the application. In domains such as medical image analysis, where the accurate detection of small, critical regions is paramount, macro metrics are preferable due to their emphasis on rare classes. Conversely, in applications like food waste estimation, weighted metrics are more appropriate. In the current research, minor segmentation errors, such as slightly mislabeling a few rice grains or having slight boundary mismatches, should have a negligible impact on the overall estimation accuracy. Thus, relying on macro metrics, which equally treat inaccuracies by minor and major classes, could unfairly penalize the model based on trivial errors in highly underrepresented classes.

In summary, although this study employs a customized weighted loss function to enhance segmentation performance for even smaller classes, weighted evaluation metrics are prioritized for model selection and optimal epoch determination. This strategy better reflects the real-world goals and tolerance thresholds inherent to the task.

## 3. Results & Analysis
### 3.1. Actual Food Waste Quantification

This section presents the results of food waste estimation based on manually annotated segmentation masks. For a comprehensive analysis, the entire dataset, comprising both the training and test sets, was used in an integrated manner. Importantly, although all data were analyzed, the estimations were conducted separately for the pre- and post-consumption stages. By applying **Equations (1-4),** the estimation of food waste was performed based on the surface area covered by each class. **Figures (8-12)** display histograms representing the distribution of each class in the segmentation masks before and after consumption. To ensure a reliable basis for comparison between the two stages, class proportions were computed using weighted averages, offering a more robust and interpretable benchmark.

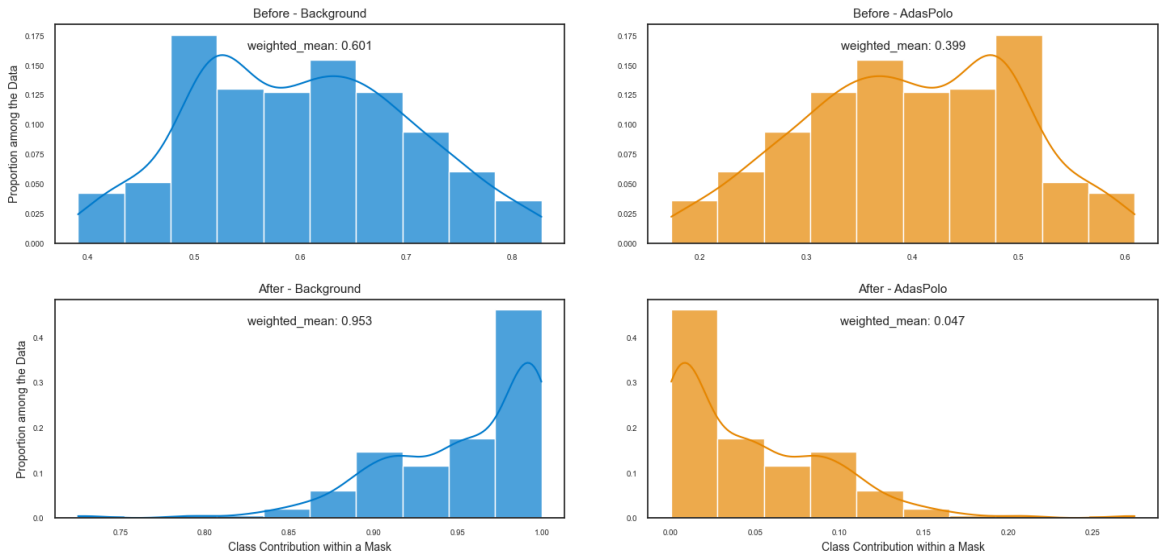

Figure 8. Histogram of Class-Wise Surface Contribution for AdasPolo Before & After Consumption

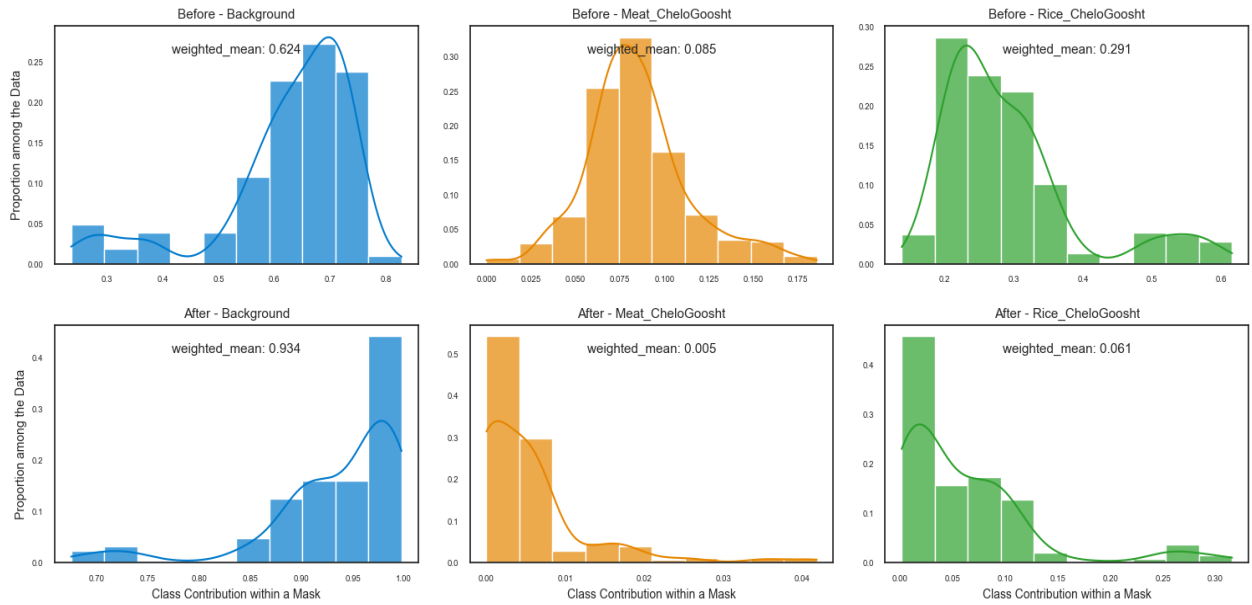

Figure 9. Histogram of Class-Wise Surface Contribution for CheloGoosht Before & After Consumption

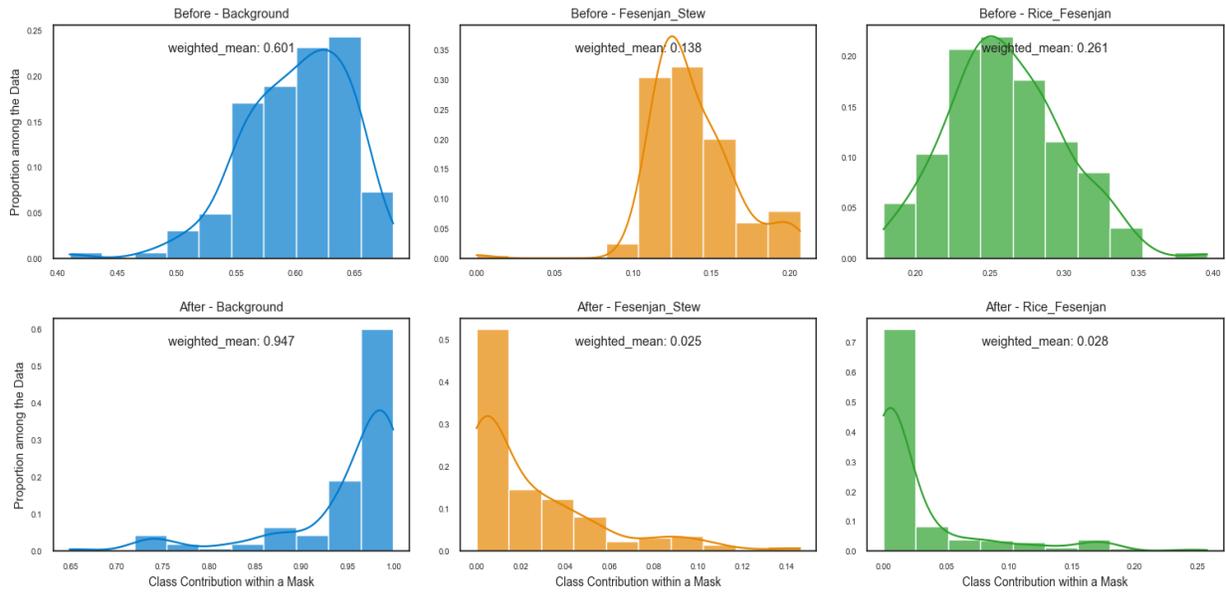

Figure 10. Histogram of Class-Wise Surface Contribution for Fesenjan Before & After Consumption

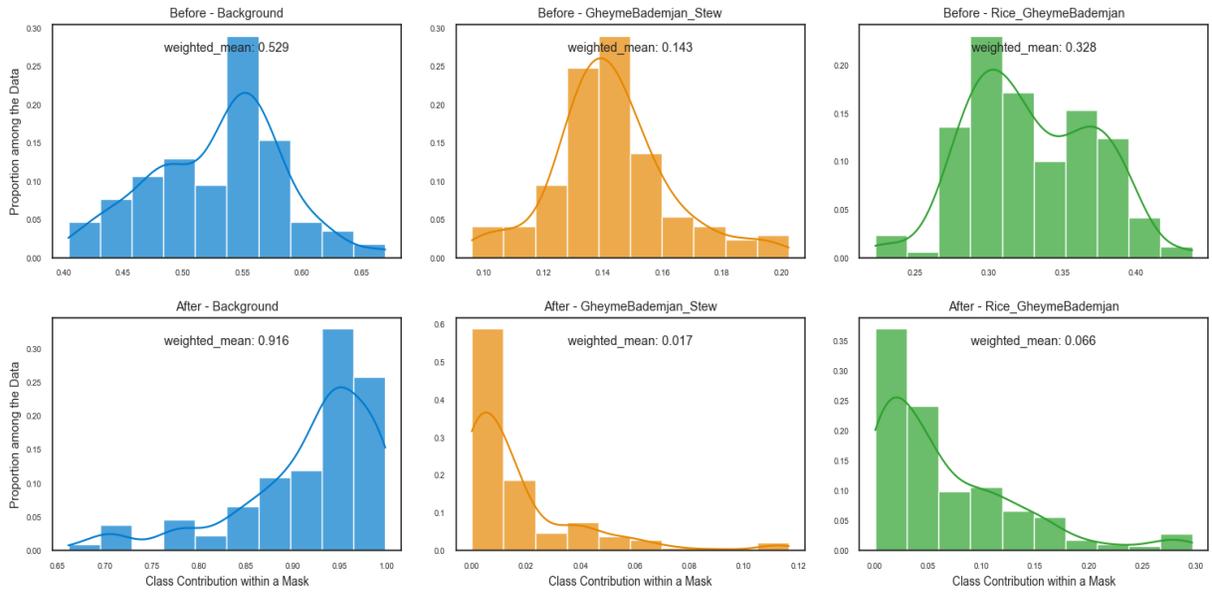

Figure 11. Histogram of Class-Wise Surface Contribution for Gheymebadmena Before & After Consumption

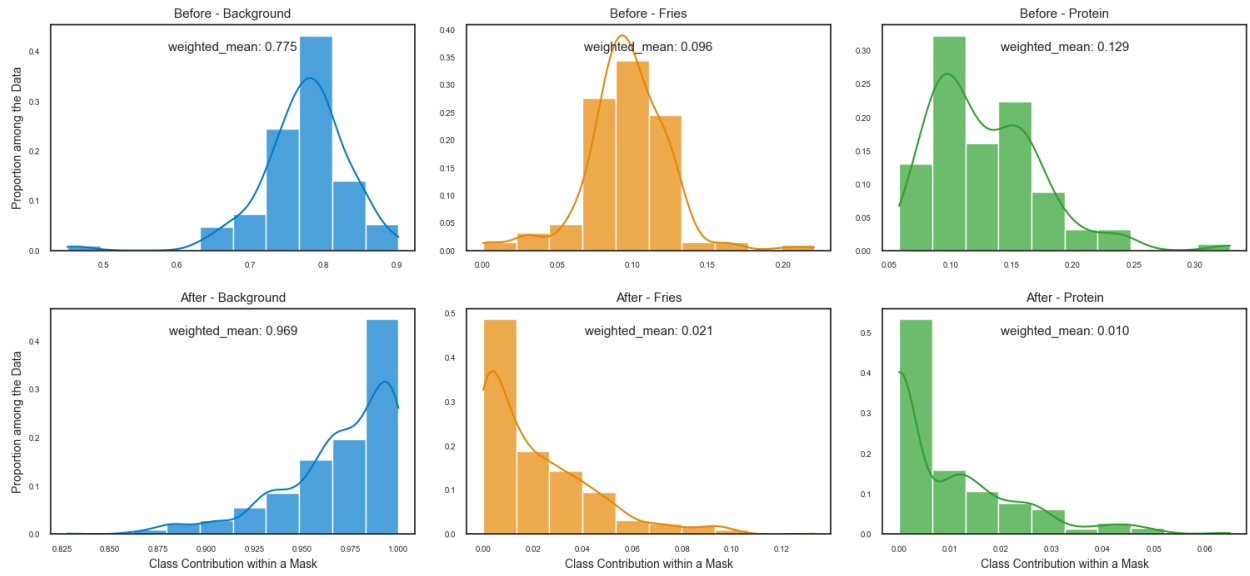

Figure 12. Histogram of Class-Wise Surface Contribution for Protein and Fries Before and After Consumption

Using the averaged values derived from the histograms and **Equations (1-4),** the mean consumption rates and corresponding food waste estimations were calculated. A detailed breakdown of these results is presented in **Table 6**.

Table 6. Average Eating and Food Waste Rates per Class Across Food Categories (Based on Surface Proportions)

| Food Type | Class | Pre-consumption Weighted Average | Post-consumption Weighted Average | Eating Rate | Remaining Rate |
|---|---|---|---|---|---|
| **AdasPolo** | 1 / AdasPolo | 0.399 | 0.047 | 88.2 | 11.8 |
| | Total | 0.399 | 0.047 | | |
| **CheloGoosht** | 1 / Meat_CehloGoosht | 0.085 | 0.005 | 94.1 | 5.9 |
| | 2 / Rice_CheloGoosht | 0.291 | 0.061 | 79.0 | 21.0 |
| | Total | 0.376 | 0.066 | | |
| **Fesenjan** | 1 / Fesenjan Stew | 0.138 | 0.025 | 81.9 | 18.1 |
| | 2 / Rice of Fesenjan | 0.261 | 0.028 | 89.3 | 10.7 |
| | Total | 0.399 | 0.053 | | |
| **GheymeBademjan** | 1 / GheymeBademjan Stew | 0.143 | 0.017 | 88.1 | 11.9 |
| | 2 / Rice of GheymeBademjan | 0.328 | 0.066 | 79.9 | 20.1 |
| | Total | 0.471 | 0.083 | | |
| **Protein & Fries** | 1 / Fries | 0.096 | 0.021 | 78.1 | 21.9 |
| | 2 / Protein | 0.129 | 0.010 | 92.2 | 7.8 |
| | Total | 0.225 | 0.031 | | |

## 3.2. Training Loop

As outlined in **Section 2.1**, the test set was separated from the beginning to ensure it remained unseen, preventing any risk of data leakage or bias. The training set was further divided into 85% for training and 15% for validation. This validation split was used to identify the best-performing model, the one achieving the highest validation weighted IoU (which is, in many cases, the highest weighted Dice as well). This model was then saved and applied to the unseen test set for final evaluation.

To provide a visual understanding of model performance over time, **Figure 13** illustrates the progression of weighted Dice scores throughout the training process for all four models across various food categories. Importantly, while some models show fluctuations during the first few epochs, early training occurs rapidly, offering a strong foundation for applying smaller learning rates in later stages to support smoother convergence. This figure also highlights the models' successful convergence toward their optimal states based on the chosen evaluation metrics.

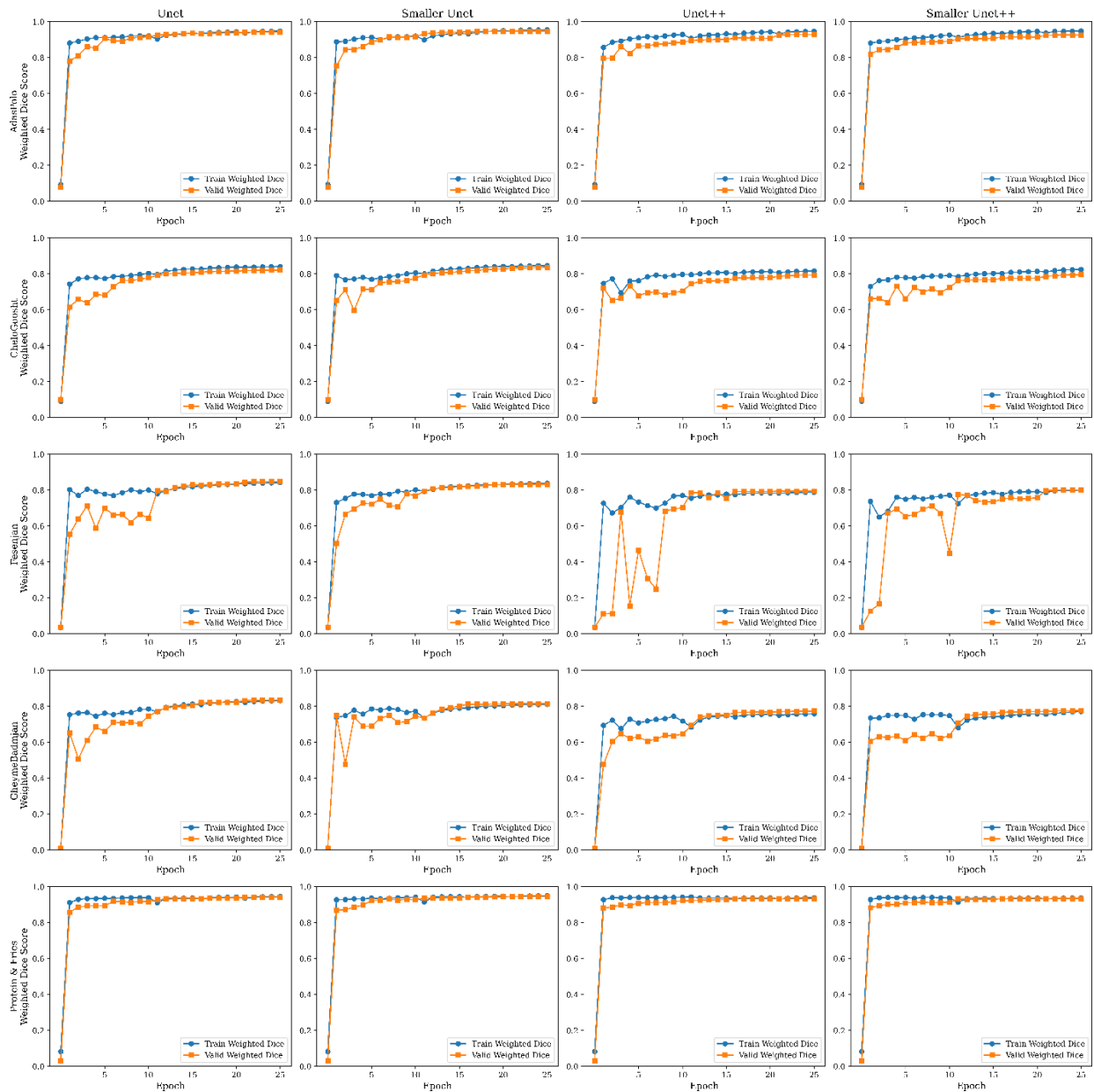

Figure 13. Training & Validation Weighted Dice Score Progression for All Models Across Food Categories

**Figure 13** further reveals that *AdasPolo* & *Protein & Fries* not only achieved the highest validation Dice scores but also demonstrated the most stable and consistent convergence patterns. Additionally, it highlights that each pair of models (Unet and Smaller Unet, as well as Unet++ and Smaller Unet++) exhibited similar performance trends and convergence behaviors when compared to models outside their respective pairs. This reflects that, at least in the current application, the model structure and inner connections may be more influential on performance compared with the number of convolutional layers, responsible for enhanced feature extraction.

Moreover, in nearly all cases, either Unet or Smaller Unet outperformed or matched the performance of Unet++ and Smaller Unet++. Notably, Smaller Unet++ also achieved results that were either better or comparable to those of the full Unet++ model. This suggests that, given the current dataset size and task complexity, the problem may not be highly challenging. In simpler terms, from a research perspective, this indicates that simpler models, with fewer parameters and a lower risk of overfitting, can perform on par with, or even surpass, more complex architectures in relatively less demanding tasks. However, we emphasize that this observation is task-specific, and, depending on the complexity and size of the dataset, testing multiple architectures remains a valid and recommended practice.

### 3.3. Performance Evaluation on Test Data

To identify the optimal model during training, the one yielding the highest validation IoU was saved and subsequently evaluated on the unseen test data, with the results summarized in **Figure 14**. This figure provides several valuable observations. In the first place, all performance metrics (particularly the DPA, as the most important one) highlight the effectiveness of the proposed approach. The maximum DPA across all models for a specific food type approached 90%, indicating the potential of these predictive models for estimating food waste with minimal reliance on manual mask annotation. Secondly, the consistently strong performance of both Unet and Smaller Unet models suggests that model selection should be aligned with task complexity. This finding underscores that more parameter-heavy models are not always superior. In scenarios such as the present study, lightweight architectures with fewer parameters and faster inference times can achieve performance comparable with or even superior to more complex counterparts.

Third, when focusing on the most stringent and widely accepted segmentation metrics (IoU and Dice) the similarity in performance between each model pair (Unet and Smaller Unet, Unet++ and Smaller Unet++) becomes evident. This observation once more confirms the earlier discussion in **Section 3.2**, reinforcing the conclusion that lighter versions of models can maintain comparable accuracy. Lastly, the categories "AdasPolo" and "Protein and Fries" consistently achieved the highest performance across all metrics, whereas the stew-based dishes (Fesenjan and GheymeBademjan) demonstrated relatively lower results. This discrepancy may stem from the inherent visual complexity and fluid nature of stews, which pose challenges for both accurate mask annotation and model prediction. This insight warrants further consideration and highlights an important aspect of dataset variability in segmentation tasks.

### 3.4. Models' Inference Speed and Computational Efficiency

To better understand how model complexity and parameter count influence computational efficiency, a detailed comparison is presented in **Figure 15.** As previously explained in **Section 2.2.3,** these measurements were obtained using a T4 GPU available on Google Colab. It is worth noting that the reported speeds may vary slightly due to factors such as internet connection and system load. Therefore, the values shown in the figure reflect the approximate minimum and maximum observed speeds, which define the performance range recorded in this study.

As anticipated, training speeds are generally lower due to the additional computational overhead of backpropagation. Moreover, model size has a clear inverse relationship with speed; models with more parameters tend to run slower. Importantly, inference speed plays a more critical role in real-world applications, where models are typically deployed for prediction. Encouragingly, all models evaluated in this study achieved inference speeds exceeding 20 images per second, supporting their potential for practical deployment. Notably, Smaller Unet, a lightweight model with approximately 7.8 million parameters, demonstrated both strong performance (as seen in **Figures 13 & 14**) and outstanding inference speed, exceeding 80 images per second. This result reinforces the idea that, depending on the task requirements and data characteristics, lightweight architectures can offer an effective trade-off between speed and performance.

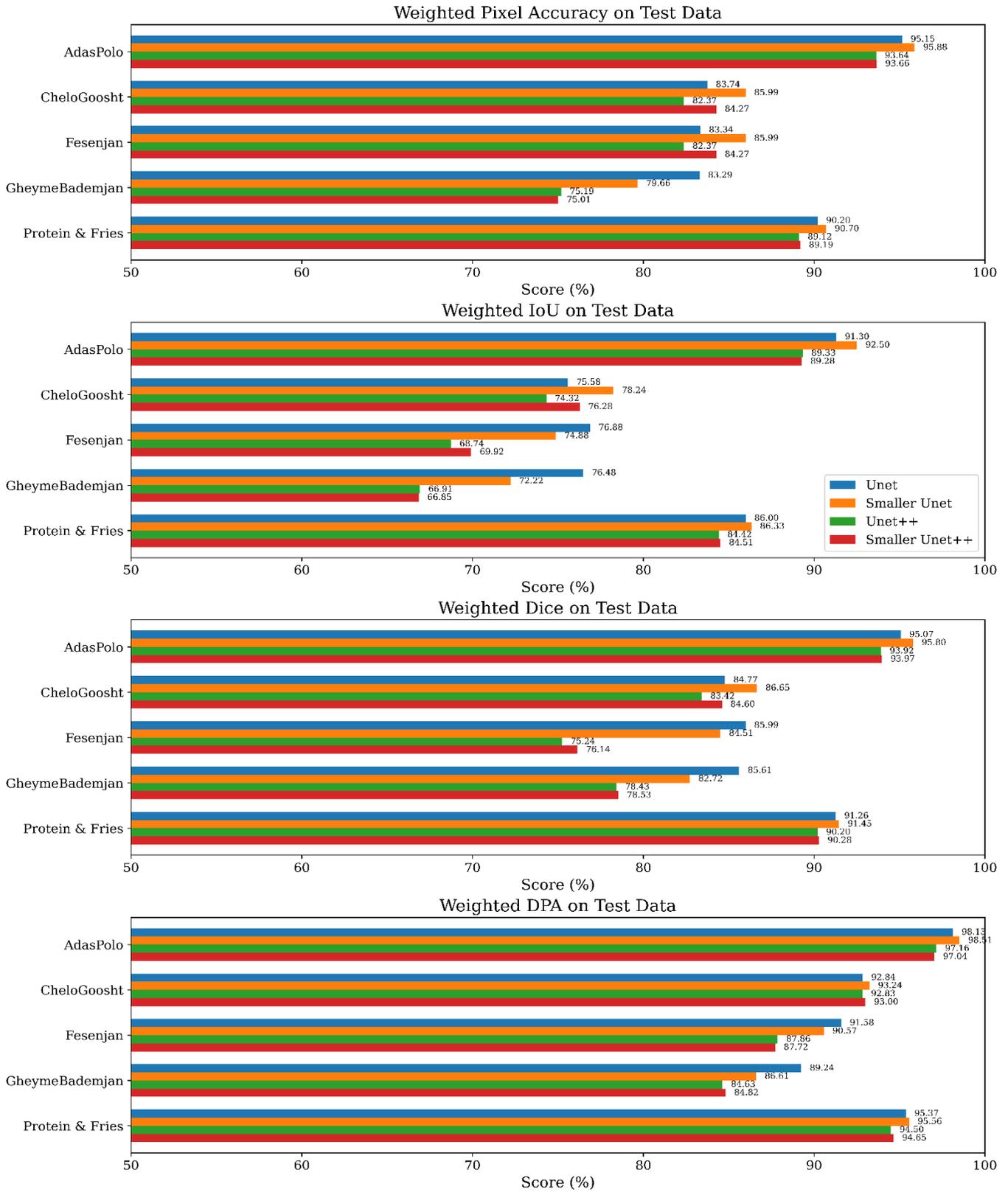

Figure 14. Comparative Performance of Segmentation Models on Test Data Across Different Food Categories

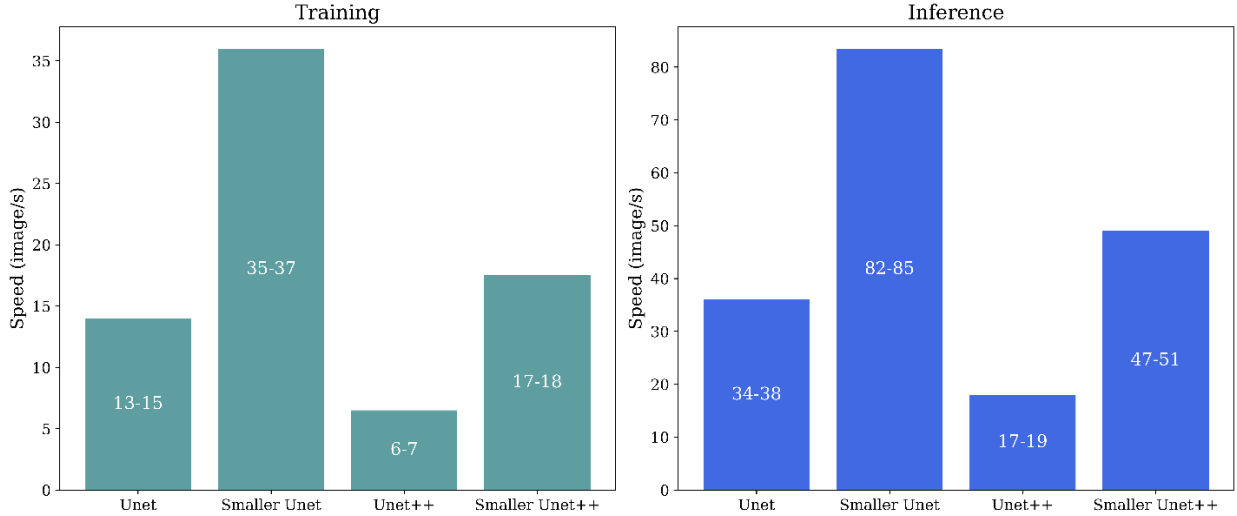

Figure 15. Training and Inference Speed Comparison

## 4. Discussion
### 4.1. Effectiveness of the Proposed Method

Undoubtedly, despite certain inherent limitations, the proposed method demonstrates strong effectiveness in addressing the task of food waste segmentation and subsequent estimation. Leveraging a predictive modeling framework enhanced with a customized loss function and carefully optimized training strategies, we achieved a stable training process and high-performing segmentation outcomes. Performance was rigorously assessed using a comprehensive suite of established metrics (Pixel Accuracy, Intersection over Union, and Dice Score), alongside our proposed metric (DPA), which prioritizes pixel class frequencies and proportions over spatial alignment, better reflecting the nature of this task. While traditional metrics remain relevant, the unique aim of this study emphasizes proportional consistency rather than pixel-level precision. As detailed in **Section 2.3.2**, the objective is not merely to achieve structurally perfect segmentations but to ensure that the class-wise pixel distributions before and after food consumption accurately mirror real-world conditions. In this context, DPA emerges as the most representative metric, capturing the degree of alignment between predicted segmentations and ground truth based on pixel frequency histograms. Nonetheless, conventional optimization targets, such as minimizing loss and maximizing Dice/IoU, remained central during training.

Encouragingly, high DPA values, mostly exceeding 0.9 in models such as Unet and Smaller Unet, demonstrate strong proportional agreement with manually annotated masks, thereby supporting the accuracy of downstream estimates like consumption and waste rates. This alignment reinforces the internal consistency of our experimental pipeline, with DPA trends supporting the quantitative outputs presented in **Sections 3.1–3.3**. Moreover, the ability of lightweight models like Smaller Unet to achieve high segmentation quality while maintaining low parameter counts, reduced FLOPs, and significantly faster inference times (see Figure 15) highlights the practicality of our approach. This is particularly valuable for real-time applications, where efficiency is essential. That said, model selection should remain task-specific; in more complex settings or with larger datasets, deeper architectures such as Unet++ may prove more suitable.

### 4.2. Considerations & Limitations

While this study advances the quantification and estimation of food waste through computer vision and semantic segmentation, it is not without limitations and challenges. This section delves into this part, improving the quality of the paper.

#### 4.2.1. 2 VS 3-Dimensional analysis

One of the inherent limitations in food waste estimation based on two-dimensional image segmentation is the absence of depth information, which impairs the model's ability to capture volumetric variations, density disparities, and overlapping food layers. For instance, in meals containing granular components such as rice, visually analogous regions in pre- and post-consumption 2D images may represent markedly different quantities due to stacking effects, pre-consumption images often depict higher volume simply due to layered presentation. Such discrepancies can introduce moderate inaccuracies in estimating consumed or remaining food volume, particularly in applications where volumetric precision is critical.

In contrast, 3D segmentation methods, which incorporate depth information (e.g., through point clouds or voxel representations), can offer a more faithful characterization of spatial and volumetric properties. This added dimension facilitates improved accuracy in quantifying food waste. Nevertheless, the practical adoption of 3D techniques remains highly limited due to significant computational, logistical, and resource-related barriers. Processing 3D data requires extensive hardware capabilities, substantially longer runtimes, and memory consumption often 10–100× higher than 2D methods, depending on the depth resolution. Additionally, generating 3D annotations, such as voxel-level masks or depth-aligned labels, is notoriously labor-intensive and costly, rendering it impractical for many research and real-world scenarios outside specialized domains like medical imaging or industrial inspection. Given these constraints, the use of 2D semantic segmentation presents itself as a pragmatic and resource-efficient alternative that balances feasibility with performance. Nonetheless, future research could explore hybrid approaches, such as the simulation or estimation of depth from 2D images, to bridge the gap between dimensional simplicity and volumetric accuracy, thus enhancing the reliability of food waste estimation.

### 4.2.2. Overestimation Risks

It is important to acknowledge that, as discussed in **Section 3.1**, the computation of consumption and remaining rates based on manual segmentation masks may exhibit a tendency toward slight underestimation of consumption and corresponding overestimation of remaining quantities. This bias is partially attributable to the factors outlined in **Section 4.2.1,** particularly the varying volumetric representation assigned to individual pixels in pre- and post-consumption states. For instance, in the pre-consumption phase, a single pixel (especially when located centrally, where food is often densely stacked) can correspond to multiple grains of rice. In contrast, the same pixel position in the post-consumption image, particularly when a significant portion has been eaten, may reflect only a single remaining grain or even partial residue. Furthermore, the manual annotation strategy aimed at generating precise ground truths for supervised model training involved labeling even the most marginal or negligible pixel regions. While this meticulous approach enhances training quality, it may inadvertently introduce minor deviations in consumption estimates, with a slight tendency to report higher consumption and lower remaining rates than might be observed through volumetrically calibrated measurements. These subtle discrepancies underscore the need for future research to explore potential corrective mechanisms, such as damping factors or calibration heuristics, that could further align pixel-based estimations with real-world volumetric changes, thereby enhancing the realism and interpretability of food consumption modeling.

### 4.2.3. Challenges in Segmenting Stew and Semi-Liquid Foods

One of the key challenges in food segmentation, particularly in post-consumption contexts, arises when dealing with foods that exhibit stew-like or semi-liquid properties. Due to their thick yet fluid consistency, such foods tend to spread across the plate during consumption, often leaving behind smeared traces or scattered remnants. This behavior complicates the process of identifying precise food boundaries. Unlike solid-structured foods that typically maintain more defined contours, even when fragmented, the delineation of stew residues is notably less straightforward. In these cases, even human annotators may find it difficult to determine whether the remaining smudges or dispersed particles should be considered part of the edible content. Furthermore, the visual properties of these foods often vary in intensity, with darker, more concentrated regions near the center fading to lighter tones at the periphery. This gradient further complicates the annotation process, introducing subjectivity and ambiguity in mask creation.

This ambiguity in human-labeled training data has a direct impact on the performance of deep learning models. Predictive models that depend on such annotations struggle to cope with the uncertain and irregular boundaries presented by these food types, as reflected in the comparatively lower evaluation scores reported for them in this study. **Figure 16** presents representative examples of post-consumption trays with differing segmentation challenges, visually highlighting the contrast between ambiguous and well-defined food remnants.

Since the root of this issue lies in the lack of a precise definition for "actual food residue" after consumption, there is a need for techniques capable of handling visual uncertainty. One promising approach is the use of Type-2 Fuzzy Logic, which has shown potential in modeling higher-order uncertainties (Castillo *et al.*, 2017), particularly in visually ambiguous regions. However, further empirical validation is needed in the context of food segmentation. Additionally, although attention mechanisms cannot resolve the ambiguity inherent in manual mask creation, incorporating them into segmentation models may still improve performance by helping the network focus more effectively on salient image regions.

### 4.2.4. Edible VS Non-edible Wastes

In this study, nearly all food items served were fully edible, eliminating the need to distinguish between edible and non-edible components in the post-consumption segmentation phase. Nevertheless, this point warrants acknowledgment. In practical scenarios where non-edible elements such as meat or poultry bones may remain after consumption, it becomes essential to assign a separate class to these items. Moreover, they should be excluded from calculations related to food intake and consumption estimation to maintain analytical accuracy.

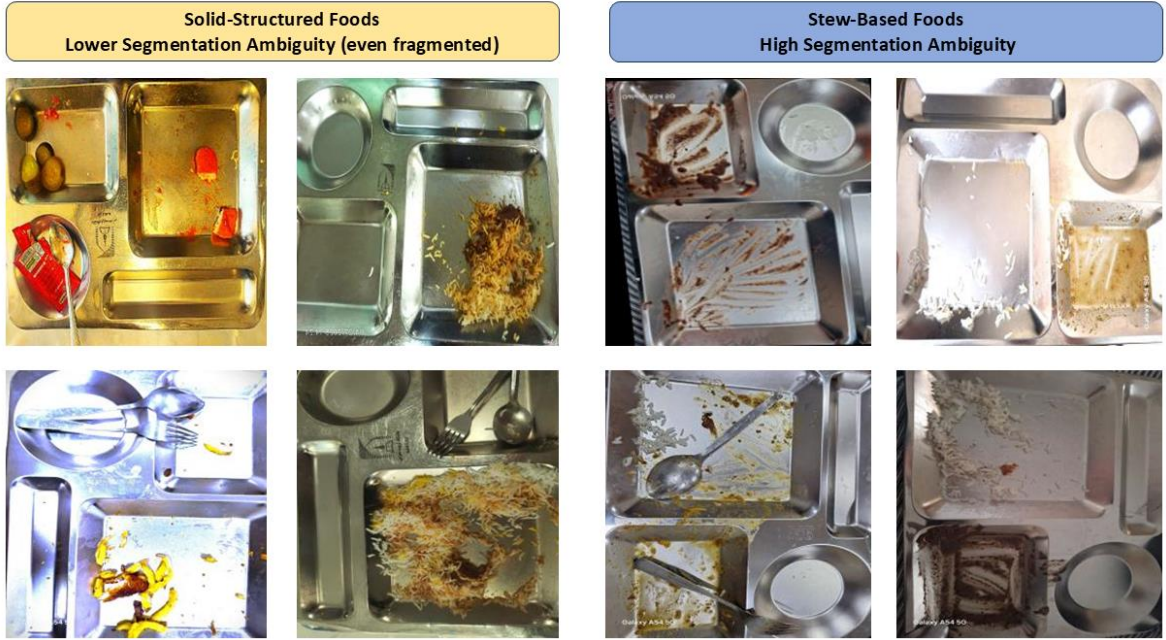

Figure 16. Examples of Post-Consumption Food Trays Illustrating Segmentation Ambiguity: Comparison Between Solid-Structured and Semi-Liquid Foods

### 4.2.5. Imaging System

The images utilized in this study were captured manually, with efforts made to maintain a relatively consistent distance across all samples. Due to budget limitations, professional-grade cameras were not accessible. However, for enhanced precision, improved consistency, and potential commercialization of the system, employing a fixed camera installed at the serving station is strongly recommended in future implementations.

## 4.3. Policy Implications

The methodology adopted in this study can have notable implications for educational institutions, environmental policymakers, and food service managers. Despite all the mentioned limitations, leveraging the proposed framework based on computer vision and deep learning still enables automated and relatively accurate monitoring of food consumption and waste, delivered in a cost-effective, fast, and non-intrusive manner. This advantage is particularly valuable in high-volume settings such as university dining halls or hospitals, where manual monitoring is difficult and inefficient. From a policy standpoint, the system designed based on the proposed method of this study holds the potential to generate actionable managerial insights, even under existing constraints. Given access to a reasonably sufficient volume of high-quality segmentation masks (the more, the better), the methodology can be trained to predict leftover quantities across different food components. For instance, our findings showed that when protein is served alongside rice (e.g., CheloGoosht) or fries (e.g., Protein & Fries), its consumption rate is notably higher. Such observations could help dining managers adjust serving portions more efficiently to minimize waste.

Additionally, the proposed method demonstrated that the suggested predictive model can operate efficiently even on lightweight architectures and with the use of mid-range hardware, supporting real-world deployment without reliance on expensive or specialized equipment. This opens a practical pathway for implementing smart monitoring systems in resource-constrained environments such as schools, dormitories, or military facilities. Finally, the presented framework may serve as a foundation for long-term behavioral analysis and food consumption monitoring, offering new avenues for future research in nutrition science, data-driven food service optimization, and industrial engineering. More studies should focus on incorporating larger and more diverse data for enhanced performance and generalization.

To sum up, while this study is subject to certain limitations—including a relatively small dataset, absence of a fixed imaging setup, and reliance on two-dimensional surface analysis—it nevertheless offers a practical, scalable, and cost-efficient framework for intelligent food waste monitoring. It is essential to note that the current research is academic in nature and is not intended for direct commercial deployment. Rather, it serves as a foundational step toward future investigations, potential real-world adaptations, and further enhancements aligned with practical business applications.

## 4.4. Future Work Direction

### 4.4.1. 3D Analysis

While two-dimensional methods have gained significantly more popularity in food waste estimation due to their simplicity and substantially higher efficiency, three-dimensional analysis still remains one of the most accurate options for evaluating the actual volume of food. Utilizing 3D data, such as point clouds or voxel-based volume segmentation, especially for foods with non-uniform depth distributions, enables a more realistic estimation of the remaining volume. However, there are considerable challenges associated with 3D processing, including the need for advanced imaging equipment, high computational load, the complexity of generating 3D masks, and substantial processing and financial costs. As a result, many researchers are compelled to avoid such approaches. Nevertheless, if proper infrastructure and financial support are available, incorporating 3D analysis in future studies could be a significant step toward reducing bias and enhancing the accuracy of food waste estimation models.

### 4.4.2. Modified 2D Analysis

Given the technical and financial challenges associated with 3D analysis, it is strongly recommended that researchers focus on enhancing two-dimensional techniques to provide more accurate and realistic estimates of food waste, thereby avoiding the substantial costs imposed by volumetric processing. In the present study, 2D segmentation was employed to compare the surface areas of food before and after consumption, yielding compelling results. However, as previously noted, surface-based analysis has inherent limitations. For instance, in pre-consumption images, particularly in the central areas of plates where food tends to accumulate, a single pixel may represent several grains of rice or stew particles. In contrast, the same pixel in post-consumption images may correspond to only one or two remaining fragments. To compensate for this volumetric bias in 2D images, one promising approach is to combine segmentation with volume estimation based on object detection In this hybrid method, during the pre-consumption phase, bounding boxes are generated around major food regions (e.g., dense clusters of rice or stew), and differential weighting is applied to central and peripheral areas (more for inner parts and less for areas closer to the borders). These weights, which can be obtained empirically or through observation, serve as estimated volumetric coefficients that can help reconstruct a more accurate approximation of true food volume and reduce the bias caused by the lack of depth information.

Additionally, one proposed strategy to further enhance the accuracy of 2D-based frameworks is to integrate them with mass-based food analysis. In this approach, surface-based segmentation is first used to estimate the approximate percentage of food consumed and the relative unconsumed portions of different food components. Subsequently, the actual weight difference between pre- and post-consumption plates is measured. This mass differential can then be employed to recalibrate the segmentation-based estimates by assigning class-specific correction factors, aligning visual predictions with physical reality. This hybrid methodology provides a more precise and cost-effective alternative to full 3D modeling, maintaining computational feasibility while enhancing volumetric inference.

### 4.4.3. Fuzzy Type-2 for Improved Segmentation

Accurate segmentation of stew-like and semi-liquid foods (especially in post-consumption states) presents persistent challenges due to ambiguous boundaries, irregular residue patterns, and highly variable color intensities. These characteristics not only make manual annotation difficult for human operators (where subjectivity and inconsistency are to some extent inevitable) but also result in training data with inherent uncertainty, which in turn complicates the task for predictive segmentation models.

Fuzzy Logic can be a promising solution in this regard. Fuzzy logic is widely recognized as one of the most effective frameworks for representing and managing uncertainty, with applications across diverse domains such as natural language processing (Liu *et al.*, 2024; Rokhva *et al.*, 2025) and computer vision (Baimukashev *et al.*, 2024; Sobrevilla and Montseny, 2003). More specifically, Type-2 Fuzzy Logic, which incorporates uncertainty directly within its membership functions, provides enhanced capability for reasoning under ambiguity. Owing to its robustness in uncertain environments, this technique has already proven valuable in fields such as medical imaging, where similar challenges regarding uncertainties are prevalent (Bose *et al.*, 2024; Castillo *et al.*, 2017, p. 2).

In the context of food segmentation, particularly for fluid or semi-fluid dishes, Type-2 fuzzy systems could be integrated into either the pre-processing or post-processing stages to better model gradual intensity variations and fuzzy contours. Unlike crisp class assignments, this approach enables the assignment of graded membership degrees to each pixel, allowing for more nuanced and flexible decision-making in scenarios with indistinct edges and color gradients. Thus, incorporating Type-2 Fuzzy Logic into future segmentation pipelines may significantly improve the handling of ambiguity and yield more reliable predictions in food waste estimation tasks.

### 4.4.4. Attention Mechanism for Improved Performance

Attention modules have emerged as one of the most effective mechanisms in enhancing the performance of deep learning models across various tasks, including classification, detection, and particularly segmentation. In the context of image segmentation, these modules enable models to focus on most significant regions while suppressing irrelevant or low-importance areas, an advantage especially valuable for food items with complex structures or irregular surface patterns. Channel Attention enhances semantically meaningful features, whereas Spatial Attention emphasizes specific spatial locations within the image. When combined, as in Convolutional Block Attention Module (CBAM), they offer a complementary mechanism for improved representation. While the baseline models used in this study, such as U-Net and U-Net++, do not incorporate attention, future extensions that integrate such modules into the encoder, bottleneck, or decoder components could lead to more precise boundary detection, better delineation of fine-grained food components, and reduced visual noise. Notably, these improvements should come with minimal additional computational cost and negligible increase in model parameters.

### 4.4.5. Extensions & Scalability

Hardware and financial constraints have been among the primary reasons for limiting the scale of the present study. Despite the wide variety of meals served in the university dining hall, this research focused on only five frequently occurring food categories and a limited number of data to ensure that manual annotation, data augmentation, and model training remained feasible within the available timeframe and resources. Naturally, this selective scope may have influenced the results. For instance, the relatively better performance of simpler models (such as the Smaller U-Net) is likely attributed to the lower data complexity and the limited range of food classes.

Nevertheless, scaling this research to real-world environments, such as commercial kitchen systems, hospitals, or military facilities, requires increased food diversity, a larger volume of annotated images, and the development of automated data collection platforms. Without these enhancements, the models are at risk of bias and may underperform in settings where food appearance varies significantly. Importantly, even in practical deployments, large, general-purpose models may not deliver optimal results. Therefore, it is recommended that, after pretraining on large-scale, diverse datasets, these models be fine-tuned using domain-specific data tailored to the particular environment (e.g., the unique menu of a specific dining hall). This dual approach, broad pretraining followed by localized adaptation, can help maintain generalizability while significantly improving model performance under real-world conditions.

## 5. Conclusion

This study introduced a novel AI- and vision-based framework for food waste estimation, specifically tailored for high-volume settings such as university dining halls. Unlike prior studies that primarily relied on manual procedures or coarse classification and detection methods, the proposed pipeline leverages precise pixel-level semantic segmentation to estimate the consumption and remaining ratios of distinct food components by globally comparing before-and-after tray images. By analyzing surface-level features in these images, the framework provides a scalable, contact-free, and cost-effective solution for approximate food waste monitoring. The core contribution lies in the design and evaluation of a 2D surface-based segmentation pipeline incorporating four customized U-Net variants, including two lightweight versions, to segment edible items such as rice, stews, and proteins. The study also integrates a tailored class-weighted loss function with capped dynamic reweighting, specifically devised to address intra-class imbalance and enhance training stability. Both standard performance metrics (e.g., Pixel Accuracy, Dice, and IoU) and a newly introduced DPA metric—designed to assess consumption distribution with an emphasis on proportional accuracy rather than strict spatial alignment—were employed for comprehensive model evaluation. Experimental results demonstrate that nearly all models achieved high performance. Notably, for each food category, at least one model reached or exceeded 90% DPA alignment. This confirms the models' effectiveness in estimating relative proportions of consumed versus remaining food, highlighting their applicability for the intended use. Interestingly, several lightweight variants matched or slightly surpassed their larger counterparts, showing that simpler architectures can offer competitive accuracy with faster inference speeds, crucial for real-time deployment. Nonetheless, as noted, these results are task-specific, and more diverse or complex data may affect performance, warranting further research. The study also acknowledges key limitations, including reliance on 2D data, limited food class diversity, and the cost of deploying fixed-angle imaging systems. Potential enhancements such as the integration of attention mechanisms, hybrid volume–mass estimation, and fuzzy logic to better address visual uncertainty in complex food textures are discussed. Despite these challenges and significant room for improvement, the proposed framework offers a pioneering foundation for intelligent, data-driven food waste estimation, with significant implications for sustainability initiatives, nutritional assessments, and data-informed policymaking in institutional dining contexts.

## Funding

This research did not receive specific grants from funding agencies in the public, commercial, or not-for-profit sectors.

## CRediT authorship contribution statement

**Shayan Rokhva:** Writing – review & editing, Writing – original draft, Visualization, Validation, Software, Resources, Project administration, Methodology, Investigation, Formal analysis, Data curation, Conceptualization. **Babak Teimourpour*:** Writing – review & editing, Validation, Supervision, Project administration.

## Declaration of Generative AI and AI-assisted technologies in the writing process

During the preparation of this work, the authors used "ChatGPT" in order to enhance language clarity and readability. After using this service, the authors reviewed and edited the content as needed and take full responsibility for the content of the publication.

## Declaration of Competing Interest

The authors declare that they have no known competing financial interests or personal relationships that could have appeared to influence the work reported in this paper.

# REFERENCES

Abiyev, R. and Adepoju, J. (2024), "Automatic Food Recognition Using Deep Convolutional Neural Networks with Self-attention Mechanism", *Human-Centric Intelligent Systems*, Vol. 4 No. 1, pp. 171–186, doi: 10.1007/s44230-023-00057-9.

Aghajani, P.F., Firouz, M.S. and Bohlol, P. (2024), "Revolutionizing Mushroom Identification: Improving efficiency with ultrasound-assisted frozen sample analysis and deep learning techniques", *Journal of Agriculture and Food Research*, Elsevier, Vol. 15, p. 100946.

Ahmadzadeh, S., Ajmal, T., Ramanathan, R. and Duan, Y. (2023), "A Comprehensive Review on Food Waste Reduction Based on IoT and Big Data Technologies", *Sustainability*, MDPI, Vol. 15 No. 4, p. 3482.

Amugongo, L.M., Kriebitz, A., Boch, A. and Lütge, C. (2023), "Mobile Computer Vision-Based Applications for Food Recognition and Volume and Calorific Estimation: A Systematic Review", *Healthcare*, Multidisciplinary Digital Publishing Institute, Vol. 11 No. 1, p. 59, doi: 10.3390/healthcare11010059.

Attiq, S., Danish Habib, M., Kaur, P., Junaid Shahid Hasni, M. and Dhir, A. (2021), "Drivers of food waste reduction behaviour in the household context", *Food Quality and Preference*, Vol. 94, p. 104300, doi: 10.1016/j.foodqual.2021.104300.

Baimukashev, R., Kadyrov, S. and Turan, C. (2024), "Systematic Survey of Deep Fuzzy Computer Vision in Biomedical Research", *Fuzzy Information and Engineering*, TUP, Vol. 16 No. 3, pp. 220–243.

Bohlol, P., Hosseinpour, S. and Firouz, M.S. (2025), "Improved food recognition using a refined ResNet50 architecture with improved fully connected layers", *Current Research in Food Science*, Elsevier, Vol. 10, p. 101005.

Bose, A., Maulik, U. and Sarkar, A. (2024), "An entropy-based membership approach on type-II fuzzy set (EMT2FCM) for biomedical image segmentation", *Engineering Applications of Artificial Intelligence*, Elsevier, Vol. 127, p. 107267.

Castillo, O., Sanchez, M.A., Gonzalez, C.I. and Martinez, G.E. (2017), "Review of Recent Type-2 Fuzzy Image Processing Applications", *Information*, Multidisciplinary Digital Publishing Institute, Vol. 8 No. 3, p. 97, doi: 10.3390/info8030097.

Chakraborty, S. and Aithal, S. (2024), "AI Kitchen", *International Journal of Applied Engineering and Management Letters*, Vol. 8, pp. 128–137, doi: 10.47992/IJAEML.2581.7000.0218.

Chawla, G. and Lugosi, P. (2025), "Driving pro-environmental practice change and food waste reduction in (and around) professional kitchens: Connecting materiality and meaning", *International Journal of Gastronomy and Food Science*, Elsevier, p. 101126.

Chen, Z., Wang, J. and Wang, Y. (2025), "Enhancing Food Image Recognition by Multi-Level Fusion and the Attention Mechanism", *Foods*, Multidisciplinary Digital Publishing Institute, Vol. 14 No. 3, p. 461, doi: 10.3390/foods14030461.

Faezirad, M., Pooya, A., Naji-Azimi, Z. and Amir Haeri, M. (2021), "Preventing food waste in subsidy-based university dining systems: An artificial neural network-aided model under uncertainty", *Waste Management & Research: The Journal for a Sustainable Circular Economy*, Vol. 39 No. 8, pp. 1027–1038, doi: 10.1177/0734242X211017974.

Fang, B., Yu, J., Chen, Z., Osman, A.I., Farghali, M., Ihara, I., Hamza, E.H., *et al.* (2023), "Artificial intelligence for waste management in smart cities: a review", *Environmental Chemistry Letters*, Vol. 21 No. 4, pp. 1959–1989, doi: 10.1007/s10311-023-01604-3.

Geetha, S., Saha, J., Dasgupta, I., Bera, R., Lawal, I.A. and Kadry, S. (2022), "Design of Waste Management System Using Ensemble Neural Networks", *Designs*, Multidisciplinary Digital Publishing Institute, Vol. 6 No. 2, p. 27, doi: 10.3390/designs6020027.

Gencia, A.D. and Balan, I.M. (2024), "Reevaluating Economic Drivers of Household Food Waste: Insights, Tools, and Implications Based on European GDP Correlations", *Sustainability*, MDPI, Vol. 16 No. 16, p. 7181.

Guerra Ibarra, J.P., Cuevas de la Rosa, F.J. and Hernandez Vidales, J.R. (2025), "Evaluation of the Effectiveness of the UNet Model with Different Backbones in the Semantic Segmentation of Tomato Leaves and Fruits", *Horticulturae*, MDPI, Vol. 11 No. 5, p. 514.

Jubayer, F., Soeb, J.A., Mojumder, A.N., Paul, M.K., Barua, P., Kayshar, S., Akter, S.S., *et al.* (2021), "Detection of mold on the food surface using YOLOv5", *Current Research in Food Science*, Vol. 4, pp. 724–728, doi: 10.1016/j.crfs.2021.10.003.

Khan, A.A., Laghari, A.A. and Awan, S.A. (2021), "Machine learning in computer vision: a review", *EAI Endorsed Transactions on Scalable Information Systems*, Vol. 8 No. 32, pp. e4–e4.

Kingma, D.P. and Ba, J. (2017), "Adam: A Method for Stochastic Optimization", arXiv, 30 January, doi: 10.48550/arXiv.1412.6980.

Klingler, N. (2024), "U-Net: A Comprehensive Guide to Its Architecture and Applications", *Viso.Ai*, 23 April, available at: https://viso.ai/deep-learning/u-net-a-comprehensive-guide-to-its-architecture-and-applications/ (accessed 2 July 2025).

Koirala, A., Walsh, K.B., Wang, Z. and McCarthy, C. (2019), "Deep learning – Method overview and review of use for fruit detection and yield estimation", *Computers and Electronics in Agriculture*, Vol. 162, pp. 219–234, doi: 10.1016/j.compag.2019.04.017.

Kumar, T.B., Prashar, D., Vaidya, G., Kumar, V., Kumar, S. and Sammy, F. (2022), "A novel model to detect and classify fresh and damaged fruits to reduce food waste using a deep learning technique", *Journal of Food Quality*, Hindawi, Vol. 2022.

Leal Filho, W., Ribeiro, P.C.C., Setti, A.F.F., Azam, F.M.S., Abubakar, I.R., Castillo-Apraiz, J., Tamayo, U., *et al.* (2024), "Toward food waste reduction at universities", *Environment, Development and Sustainability*, Springer, Vol. 26 No. 7, pp. 16585–16606.

Liu, M., Zhang, H., Xu, Z. and Ding, K. (2024), "The fusion of fuzzy theories and natural language processing: A state-of-the-art survey", *Applied Soft Computing*, Elsevier, p. 111818.

Loshchilov, I. and Hutter, F. (2019), "Decoupled Weight Decay Regularization", arXiv, 4 January, doi: 10.48550/arXiv.1711.05101.

Lubura, J., Pezo, L., Sandu, M.A., Voronova, V., Donsì, F., Šic Žlabur, J., Ribić, B., *et al.* (2022), "Food Recognition and Food Waste Estimation Using Convolutional Neural Network", *Electronics*, MDPI, Vol. 11 No. 22, p. 3746.

Mazloumian, A., Rosenthal, M. and Gelke, H. (2020), "Deep Learning for Classifying Food Waste", arXiv, 6 February.

Moumane, K., El Asri, I., Cheniguer, T. and Elbiki, S. (2023), "Food Recognition and Nutrition Estimation using MobileNetV2 CNN architecture and Transfer Learning", *2023 14th International Conference on Intelligent Systems: Theories and Applications (SITA)*, presented at the 2023 14th International Conference on Intelligent Systems: Theories and Applications (SITA), pp. 1–7, doi: 10.1109/SITA60746.2023.10373725.

Mustapha, A.A., Saruchi, S. 'Atifah, Supriyono, H. and Solihin, M.I. (2025), "A Hybrid Deep Learning Model for Waste Detection and Classification Utilizing YOLOv8 and CNN", *Engineering Proceedings*, Multidisciplinary Digital Publishing Institute, Vol. 84 No. 1, p. 82, doi: 10.3390/engproc2025084082.

Onyeaka, H., Tamasiga, P., Nwauzoma, U.M., Miri, T., Juliet, U.C., Nwaiwu, O. and Akinsemolu, A.A. (2023), "Using Artificial Intelligence to Tackle Food Waste and Enhance the Circular Economy: Maximising Resource Efficiency and Minimising Environmental Impact: A Review", *Sustainability*, Multidisciplinary Digital Publishing Institute, Vol. 15 No. 13, p. 10482, doi: 10.3390/su151310482.

"Papers with Code - UNet++ Explained". (n.d.). , available at: https://paperswithcode.com/method/unet (accessed 2 July 2025).

Park, D., Lee, J., Lee, J. and Lee, K. (2021), "Deep Learning based Food Instance Segmentation using Synthetic Data", arXiv, 20 July, doi: 10.48550/arXiv.2107.07191.

Razavi, M., Mavaddati, S. and Koohi, H. (2024), "ResNet deep models and transfer learning technique for classification and quality detection of rice cultivars", *Expert Systems with Applications*, Vol. 247, p. 123276, doi: 10.1016/j.eswa.2024.123276.

Rokhva, S. and Teimourpour, B. (2025), "Accurate & real-time food classification through the synergistic integration of EfficientNetB7, CBAM, transfer learning, and data augmentation", *Food and Humanity*, Vol. 4, p. 100492, doi: 10.1016/j.foohum.2024.100492.

Rokhva, S., Teimourpour, B. and Babaei, R. (2025), "Enhanced Sentiment Analysis of Iranian Restaurant Reviews Utilizing Sentiment Intensity Analyzer & Fuzzy Logic", *Food and Humanity*, Elsevier, p. 100658.

Rokhva, S., Teimourpour, B. and Soltani, A.H. (2024a), "Computer vision in the food industry: Accurate, real-time, and automatic food recognition with pretrained MobileNetV2", *Food and Humanity*, Vol. 3, p. 100378, doi: 10.1016/j.foohum.2024.100378.

Rokhva, S., Teimourpour, B. and Soltani, A.H. (2024b), "Ai in the Food Industry: Utilizing Efficientnetb7 & Transfer Learning for Accurate and Real-Time Food Recognition", SSRN Scholarly Paper, Rochester, NY, 24 July, doi: 10.2139/ssrn.4903767.

Ronneberger, O., Fischer, P. and Brox, T. (2015), "U-Net: Convolutional Networks for Biomedical Image Segmentation", arXiv, 18 May, doi: 10.48550/arXiv.1505.04597.

Rukundo, O. (2023), "Effects of Image Size on Deep Learning", *Electronics*, Multidisciplinary Digital Publishing Institute, Vol. 12 No. 4, p. 985, doi: 10.3390/electronics12040985.

Shehzad, K., Ali, U. and Munir, A. (2025), "Computer Vision for Food Quality Assessment: Advances and Challenges", SSRN Scholarly Paper, Social Science Research Network, Rochester, NY, 1 January, doi: 10.2139/ssrn.5196776.

Sigala, E.G., Gerwin, P., Chroni, C., Abeliotis, K., Strotmann, C. and Lasaridi, K. (2025), "Reducing food waste in the HORECA sector using AI-based waste-tracking devices", *Waste Management*, Elsevier, Vol. 198, pp. 77–86.

Sobrevilla, P. and Montseny, E. (2003), "Fuzzy sets in computer vision: an overview", *Mathware and Soft Computing*, Universitat Politecnica de Catalunya; 1998, Vol. 10 No. 2/3, pp. 71–83.

Tao, L., Burghardt, T., Mirmehdi, M., Damen, D., Cooper, A., Hannuna, S., Camplani, M., *et al.* (2017), "Calorie counter: RGB-depth visual estimation of energy expenditure at home", *Computer Vision–ACCV 2016 Workshops: ACCV 2016 International Workshops, Taipei, Taiwan, November 20-24, 2016, Revised Selected Papers, Part I 13*, Springer, pp. 239–251.

Thaseen Ikram, S., Mohanraj, V., Ramachandran, S. and Balakrishnan, A. (2023), "An Intelligent Waste Management Application Using IoT and a Genetic Algorithm–Fuzzy Inference System", *Applied Sciences*, Multidisciplinary Digital Publishing Institute, Vol. 13 No. 6, p. 3943, doi: 10.3390/app13063943.

Vardopoulos, I., Abeliotis, K. and Lasaridi, K. (2025), "A Systematic Informetric Analysis and Literature Review of Food Waste Quantification Studies in the Food Service Industry", *Sustainability*, Multidisciplinary Digital Publishing Institute, Vol. 17 No. 1, p. 103, doi: 10.3390/su17010103.

Wang, W., Zhu, A., Wei, H. and Yu, L. (2024), "A novel method for vegetable and fruit classification based on using diffusion maps and machine learning", *Current Research in Food Science*, Vol. 8, p. 100737, doi: 10.1016/j.crfs.2024.100737.

Yuan, Y. and Chen, X. (2024), "Vegetable and fruit freshness detection based on deep features and principal component analysis", *Current Research in Food Science*, Vol. 8, p. 100656, doi: 10.1016/j.crfs.2023.100656.

Zhang, Y., Yang, X., Cheng, Y., Wu, X., Sun, X., Hou, R. and Wang, H. (2024), "Fruit freshness detection based on multi-task convolutional neural network", *Current Research in Food Science*, Vol. 8, p. 100733, doi: 10.1016/j.crfs.2024.100733.

Zhou, Z., Siddiquee, M.M.R., Tajbakhsh, N. and Liang, J. (2018), "UNet++: A Nested U-Net Architecture for Medical Image Segmentation", arXiv, 18 July, doi: 10.48550/arXiv.1807.10165.

Zhu, L., Spachos, P., Pensini, E. and Plataniotis, K.N. (2021), "Deep learning and machine vision for food processing: A survey", *Current Research in Food Science*, Elsevier, Vol. 4, pp. 233–249.